\documentclass{article}

\PassOptionsToPackage{numbers, compress}{natbib}

\usepackage[final]{neurips_data_2024}

\usepackage[utf8]{inputenc} 
\usepackage[T1]{fontenc}    
\usepackage[dvipsnames]{xcolor}         
\usepackage[pagebackref,breaklinks,colorlinks]{hyperref}       
 \hypersetup{
     colorlinks=true,
     linkcolor=OrangeRed,
     filecolor=blue,
     citecolor = NavyBlue,      
     urlcolor= Magenta,
     }
\usepackage{url}            
\usepackage{booktabs}       
\usepackage{amsfonts}       
\usepackage{nicefrac}       
\usepackage{microtype}      
\usepackage{amsmath}
\usepackage{algpseudocode}
\usepackage{algorithm}
\usepackage{graphicx}
\usepackage{fdsymbol}
\usepackage{multirow}
\usepackage{colortbl}
\usepackage{makecell}
\usepackage{algpseudocode}
\usepackage{pifont}
\usepackage{subcaption}
\usepackage{cleveref}
\usepackage{mwe,tikz}
\usepackage[percent]{overpic}
\usepackage[export]{adjustbox}
\newcommand{\blur}[1]{%
\reflectbox{\scalebox{-1}[1]{\includegraphics[width=9.4pt]{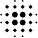}}}
}
\newcommand{\blurlow}[1]{%
\reflectbox{\scalebox{-1}[1]{\includegraphics[width=9.4pt]{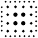}}}
}
\newcommand{\blurhigh}[1]{%
\reflectbox{\scalebox{-1}[1]{\includegraphics[width=9.4pt]{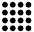}}}
}
\newcommand{\contrast}[1]{%
\reflectbox{\scalebox{-1}[1]{\includegraphics[width=9.4pt]{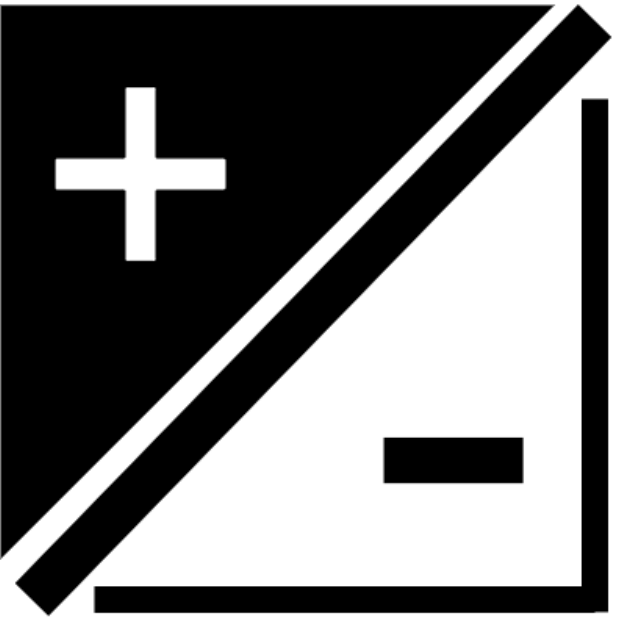}}}
}
\newcommand{\contrastlow}[1]{%
\reflectbox{\scalebox{-1}[1]{\includegraphics[width=9.4pt]{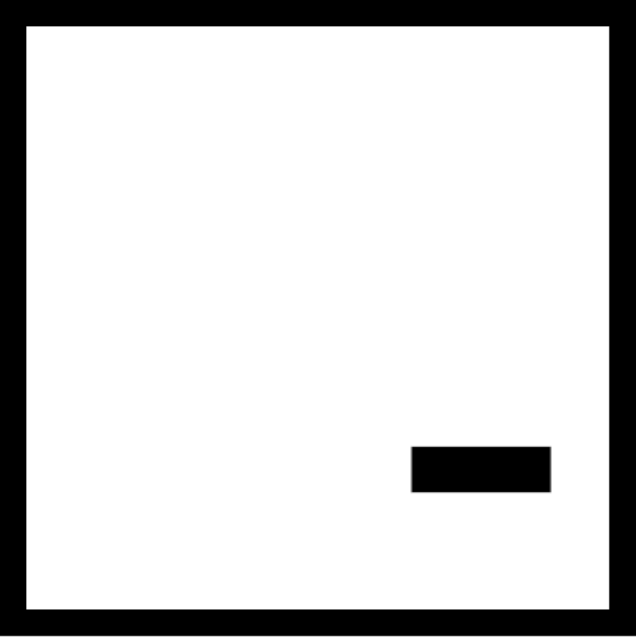}}}
}
\newcommand{\contrasthigh}[1]{%
\reflectbox{\scalebox{-1}[1]{\includegraphics[width=9.4pt]{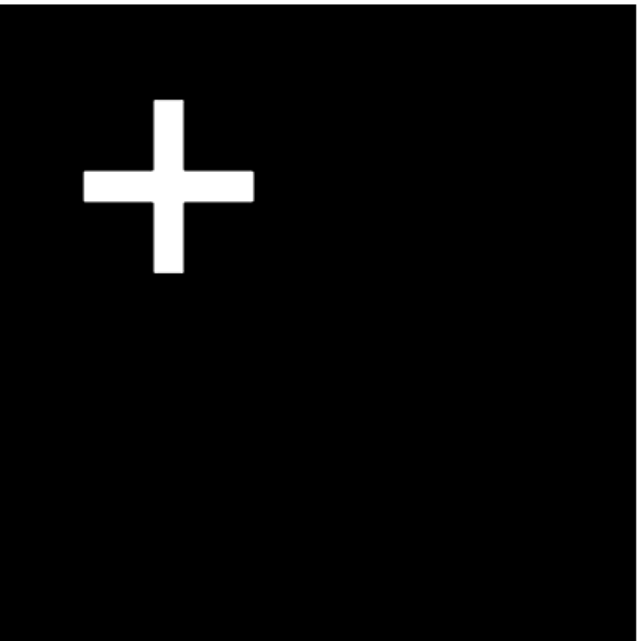}}}
}
\newcommand{\lum}[1]{%
\reflectbox{\scalebox{-1}[1]{\includegraphics[width=9.4pt]{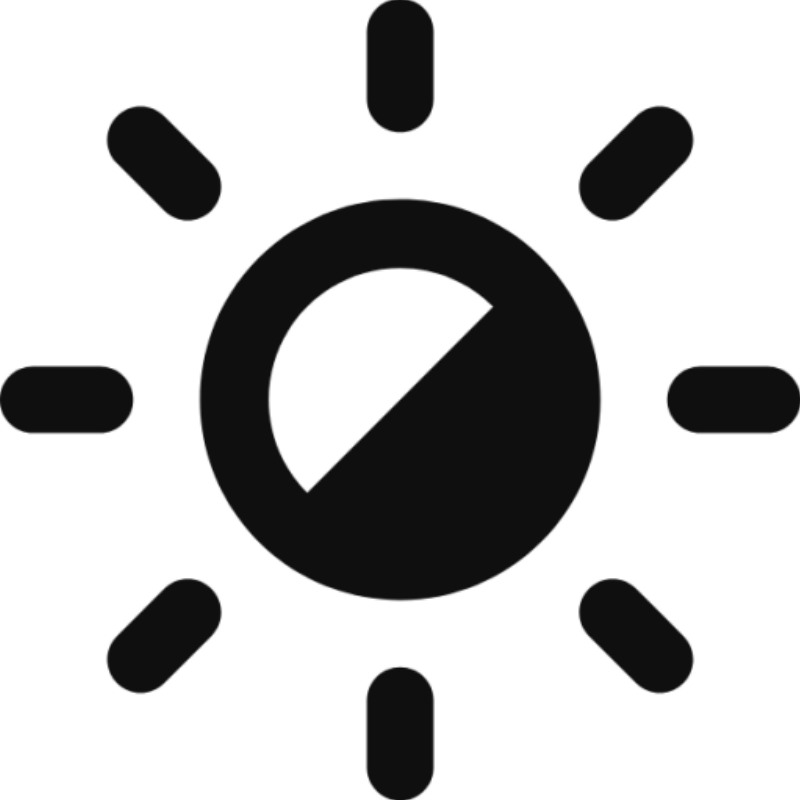}}}
}
\newcommand{\lumlow}[1]{%
\reflectbox{\scalebox{-1}[1]{\includegraphics[width=9.4pt]{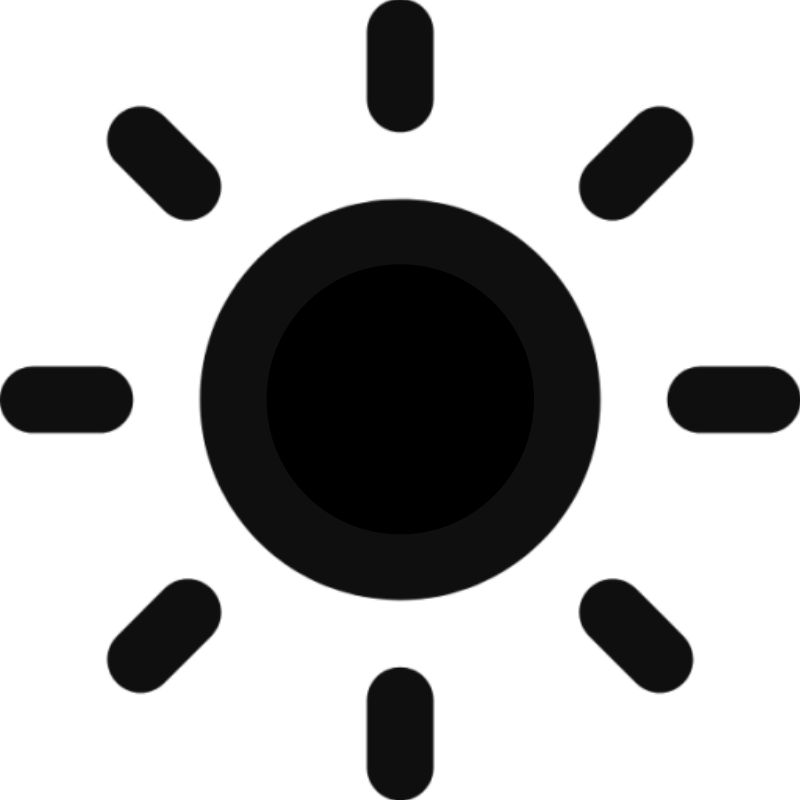}}}
}
\newcommand{\lumhigh}[1]{%
\reflectbox{\scalebox{-1}[1]{\includegraphics[width=9.4pt]{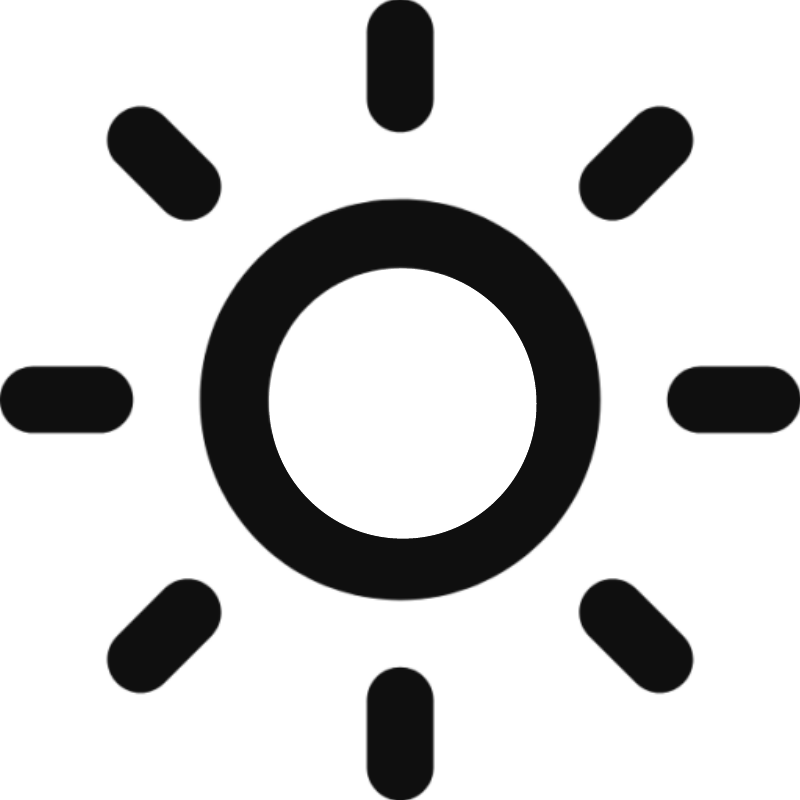}}}
}

\newcommand{\motion}[1]{%
\reflectbox{\scalebox{-1}[1]{\includegraphics[width=11pt,trim=-.2em 0 1.8em -.2em]{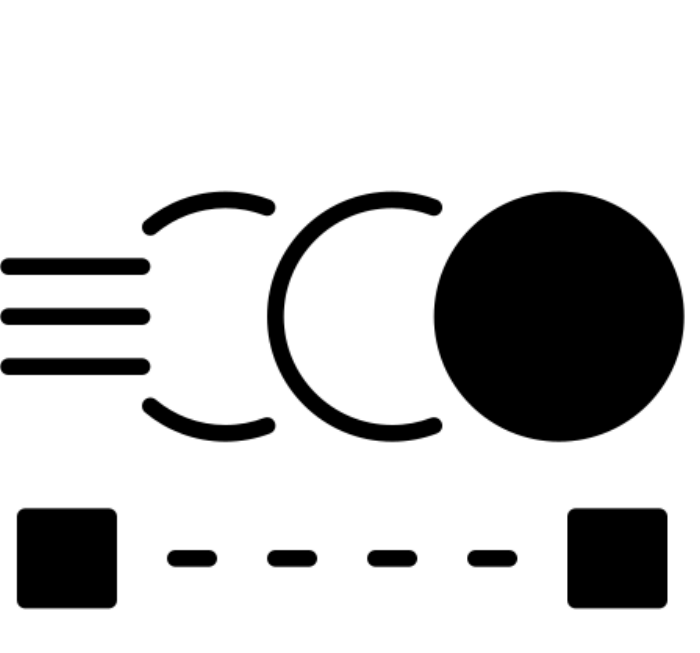}}}
}
\newcommand{\motionfast}[1]{%
\reflectbox{\scalebox{-1}[1]{\includegraphics[width=11pt,trim=-.2em 0 1.8em -.2em]{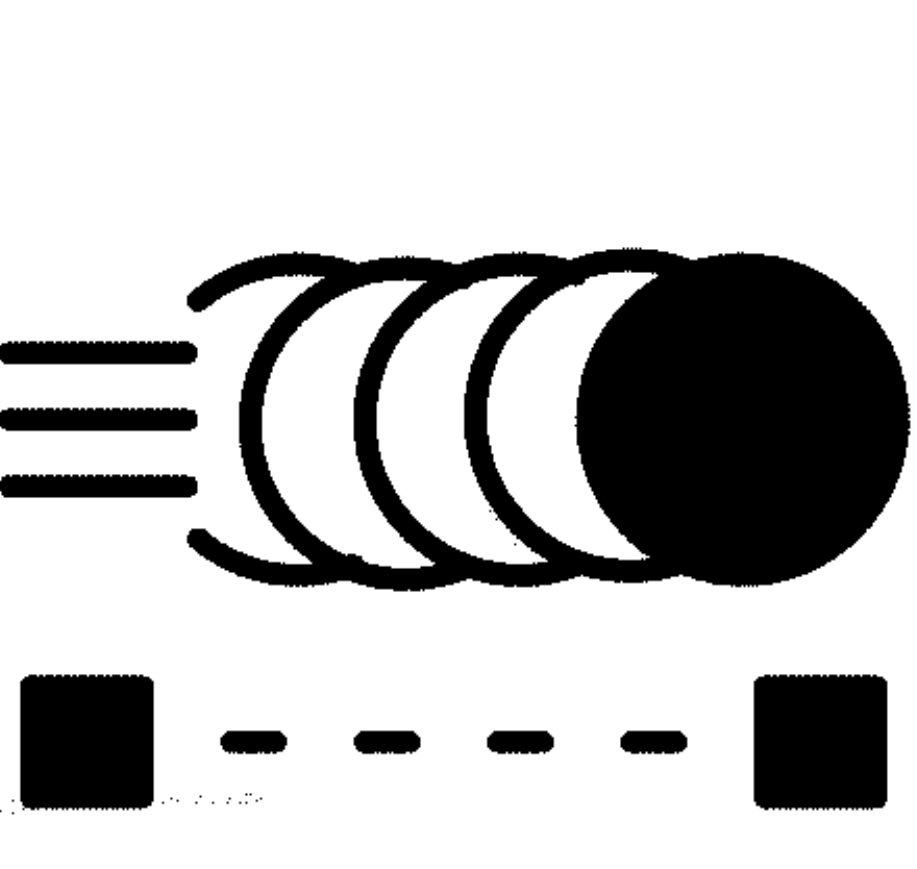}}}
}
\newcommand{\motionslow}[1]{%
\reflectbox{\scalebox{-1}[1]{\includegraphics[width=11pt,trim=-.2em 0 1.8em -.2em]{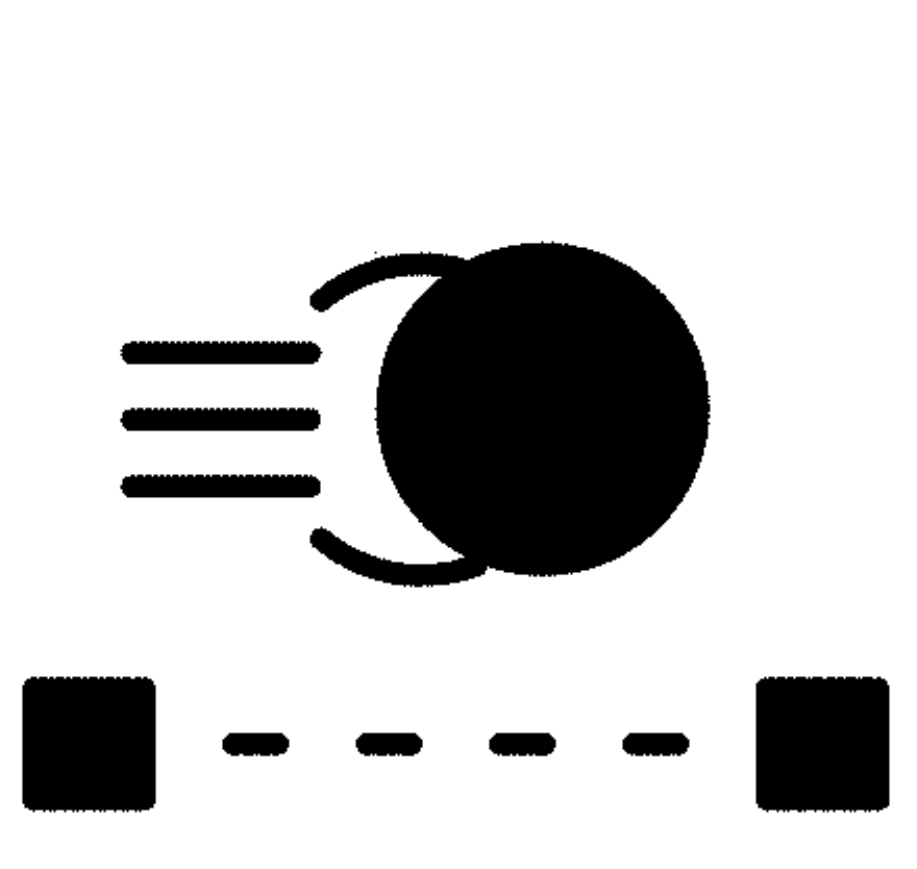}}}
}

\newcommand\tstrut{\rule{0pt}{2.4ex}}
\newcommand\bstrut{\rule[-1.0ex]{0pt}{0pt}}

\Crefname{equation}{Eq.}{Eqs.}
\Crefname{figure}{Fig.}{Figs.}
\Crefname{table}{Tab.}{Tabs.}

\definecolor{col1_}{HTML}{D2E0FB}
\definecolor{col2_}{HTML}{D4E7C5}
\definecolor{col3_}{HTML}{DDB6C6}
\definecolor{LightGrey}{rgb}{0.9,0.9,0.9}
\definecolor{col1}{HTML}{51829B}
\definecolor{col2}{HTML}{BACD92}
\definecolor{col3}{HTML}{D37676}
\definecolor{col4}{HTML}{ECA869}
\definecolor{col5}{HTML}{804674}

\definecolor{misc}{HTML}{ECA869}
\definecolor{act}{HTML}{BACD92}
\definecolor{camera}{HTML}{804674}

\definecolor{all}{HTML}{99B8FF}
\definecolor{rec}{HTML}{FA7C27}
\definecolor{control}{HTML}{B5739D}

\newcommand\Appendix{%
  \xdef\presupfigures{\arabic{figure}}
  \xdef\presupsections{\arabic{section}}
  \renewcommand\thefigure{A\fpeval{\arabic{figure}-\presupfigures}}
  \renewcommand\thesection{A\fpeval{\arabic{section}-\presupsections}}
  \renewcommand{\thetable}{A\arabic{table}}
  \renewcommand{\theequation}{A\arabic{equation}}
}

\title{LAVIB: A Large-scale Video Interpolation Benchmark}

\author{%
  Alexandros Stergiou \\
  University of Twente, NL\\
  \texttt{a.g.stergiou@utwente.nl} \\
}

\begin{document}

\maketitle

\begin{abstract}
  This paper introduces a \textbf{LA}rge-scale \textbf{V}ideo \textbf{I}nterpolation \textbf{B}enchmark (LAVIB) for the low-level video task of Video Frame Interpolation (VFI). LAVIB comprises a large collection of high-resolution videos sourced from the web through an automated pipeline with minimal requirements for human verification. Metrics are computed for each video's motion magnitudes, luminance conditions, frame sharpness, and contrast. The collection of videos and the creation of quantitative challenges based on these metrics are under-explored by current low-level video task datasets. In total, LAVIB includes 283K clips from 17K ultra-HD videos, covering 77.6 hours. Benchmark train, val, and test sets maintain similar video metric distributions. Further splits are also created for out-of-distribution (OOD) challenges, with train and test splits including videos of dissimilar attributes. \footnote{LAVIB is accessible at \url{https://alexandrosstergiou.github.io/datasets/LAVIB}.}
\end{abstract}

\section{Introduction}
\label{sec:intro}

\begin{figure}[t]
    \centering
    \begin{overpic}[width=\linewidth]{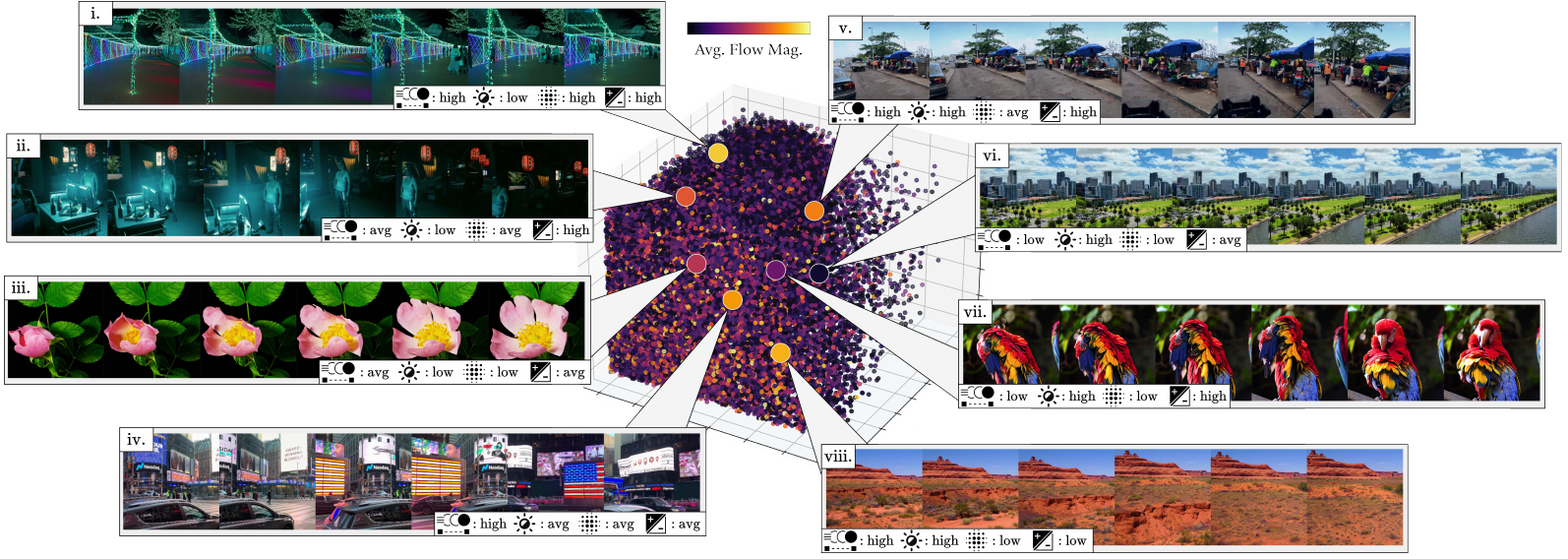}
     \put(92,0){\includegraphics[scale=0.115]{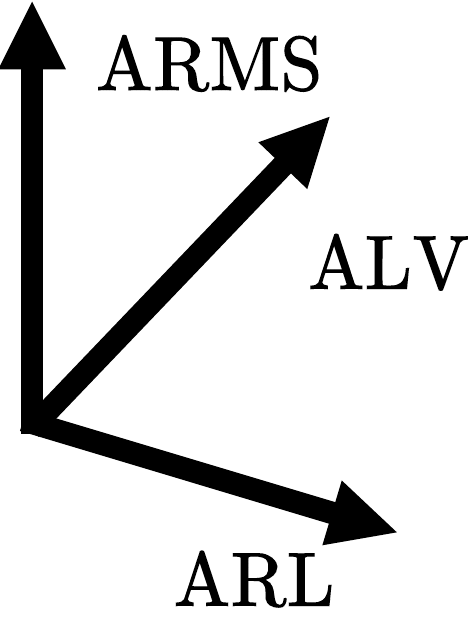}}
     \end{overpic}
    \caption{\textbf{LAVIB videos distributed across metrics}. Four metrics are computed per video. Average Flow Magnitude (AFM) quantifies motion~\motion~. The Average Laplacian Variance (ALV) is used to describe the sharpness of frames~\blur~. The Average Root Mean Square (ARMS) is used for contrast~\contrast~. The Average Relevant Luminance (ARL) relates to the video brightness~\lum~. The four aforementioned metrics are used for Out-Of-Distribution (ODD) challenges: Fast~\motionfast~~\raisebox{.2em}{$\rightarrow$} slow~\motionslow~ and slow~\motionslow~~\raisebox{.2em}{$\rightarrow$} fast~\motionfast~ motions. Low~\blurlow~~\raisebox{.2em}{$\rightarrow$} high~\blurhigh~and high~\blurhigh~~\raisebox{.2em}{$\rightarrow$} low~\blurlow~sharpness. Low~\contrastlow~~\raisebox{.2em}{$\rightarrow$} high~\contrasthigh~and high~\contrasthigh~~\raisebox{.2em}{$\rightarrow$} low~\contrastlow~contrast. Bright~\lumhigh~~\raisebox{.2em}{$\rightarrow$} dark~\lumlow~and dark~\lumlow~~\raisebox{.2em}{$\rightarrow$} bright~\lumhigh~luminance.}  
    \label{fig:teaser}
    \vspace{-1.2em}
\end{figure}

\looseness-1 Long uncompressed video streams capture events over varying motion intensities, light conditions, and color dynamic ranges. Although loading and storing individual videos is rudimentary, processing and reading large volumes can bottleneck availability. The high-volume transfer of videos with large filesizes can also result in bandwidth overheads and long decoding times. Low-level vision tasks such as Video Frame Interpolation (VFI) \cite{bao2019depth,ding2021cdfi,huang2022real,jiang2018super,jin2023unified,li2020video,niklaus2020softmax,niklaus2017video,park2021asymmetric,reda2022film,zhang2023extracting}, Video Super-Resolution (VSR)~\cite{chan2022basicvsr,hu2022you,isobe2022look,ji2020real,jo2018deep,kupyn2019deblurgan,li2021comisr,shi2022rethinking,tao2017detail,wang2020deep}, and Video Denoising (VD)~\cite{huang2022neural,liang2024vrt,liang2022recurrent,sheth2021unsupervised,song2022tempformer,su2017deep,tassano2019dvdnet} aim to address such challenges by enabling the storage and stream of lower-resolution, lower-frame-rate, compressed videos. Despite the wide application of such approaches to adjacent tasks such as localization and mapping~\cite{kim2023event,xu2021moving}, object tracking~\cite{yoo2023video}, novel view synthesis~\cite{paliwal2023implicit,shangguan2022learning}, and slow-motion video generation~\cite{jiang2018super,jin2019learning}, existing datasets for low-level video tasks~\cite{baker2011database,nah2017deep,niklaus2020softmax,perazzi2016benchmark,sim2021xvfi,siyao2021deep,soomro2012ucf101,stergiou2022adapool,su2017deep,xue2019video} contain short videos, with a small number of frames per video. With the exception of~\cite{xue2019video} most of these datasets only include either a few hundreds~\cite{baker2011database,niklaus2020softmax,perazzi2016benchmark,su2017deep} or thousands~\cite{nah2017deep,sim2021xvfi,siyao2021deep,soomro2012ucf101,stergiou2022adapool} of videos with limited variations in the motions, luminance, and object-level sharpness. To address this gap, this paper introduces a \textbf{LA}rge \textbf{V}ideo \textbf{I}nterpolation \textbf{B}enchmark (\textbf{LAVIB}), for learning to interpolate high-resolution videos across varying motion, blur, luminance, and contrast settings. LAVIB is built on per-frame metrics that quantitatively measure motion magnitudes, frame sharpness, video contrast, and overall luminance. In~\Cref{fig:teaser}, LAVIB videos are visualized over axes corresponding to the metrics used. 

The selected metrics establish a diverse, general, and robust benchmark for VFI as most prior efforts have focused on specific settings. Seminal works~\cite{montgomery1994xiph,sim2021xvfi} sourced videos from high frame-rate sensors that are less relevant to videos recorded by commonly used devices. Other works use videos of standardized resolutions and frame rates. These are either datasets of larger sizes with low-resolution videos~\cite{soomro2012ucf101,xue2019video} or smaller datasets of high-resolution~\cite{niklaus2020softmax,sim2021xvfi,siyao2021deep,stergiou2022adapool,su2017deep}. Comparisons to other video datasets across metrics are discussed in \S\ref{sec:related}.  

LAVIB contains 283,484 video segments totaling approximately 77.6 hours. The segments are sourced from 17,204 clips with $3840 \times 2160$ (4K) resolution and 60 frames-per-second (fps). Statistics are discussed in \S\ref{sec:statistics}. Similar to previous efforts comprised of 4K videos~\cite{montgomery1994xiph,sim2021xvfi,stergiou2022adapool}, LAVIB is compiled by temporally and spatially cropping tubelets from the 4K videos to fit clips into memory. 

\looseness-1 The data collection pipeline for LAVIB is detailed in \S\ref{sec:annotation}. This includes the creation of a vocabulary of search query terms. Clips are sourced from YouTube videos queried by search terms. Preset clip sampling intervals are used to standardize durations. Segments are selected from high average flow magnitude temporal locations calculated with~\cite{huang2022flowformer}. Spatial locations are selected by tubelets of high/low metrics values. The final train/val/test sets are constructed by balancing all metrics. 

\looseness-1 Widely-used VFI methods~\cite{huang2022real,kalluri2023flavr,zhang2023extracting} are benchmarked on the LAVIB val and test sets in \S\ref{sec:results}. Performance is reported across well-adopted evaluation metrics~\cite{bovik2017robust,czolbe2020loss,ding2020image,hou2022perceptual,li2019quality,zhang2018unreasonable}. LAVIB's large size and video diversity enables pretraining models of greater generalizability that are in turn evaluated on test sets of smaller down-stream datasets targeting either scene diversity~\cite{xue2019video}, high frame rates~\cite{sim2021xvfi}, or high video resolution~\cite{montgomery1994xiph,niklaus2020softmax}. In addition to the main benchmark splits, four challenges with two settings each, are introduced for Out-Of-Distribution (OOD) VFI. Train, val, and test sets with unbalanced metric distributions are created for each challenge and setting. Videos are assigned to sets based on their average motion magnitude, sharpness, contrast, and luminance metrics. These challenges evaluate model generalizability over diverse domains that are different in the train and test sets.

\section{Related works}
\label{sec:related}


Initial VFI benchmarks~\cite{baker2011database} provided real image sequences and ground truth optical flow annotations with average resolutions of $640 \times 480$. The dataset comprised a small number of videos used primarily for evaluation. Vid4~\cite{liu2013bayesian} is a standardized testing benchmark for VFI and VSR consisting of four videos of $740 \times 480$ and $720 \times 576$ resolutions. Similarly,~\cite{yi2019progressive} is also used for VSR with videos sampled from~\cite{wang2018multi}. Later efforts~\cite{nah2019ntire} have also introduced benchmarks for VD in tandem with VSR and VFI. More recent works~\cite{nah2017deep} included 3.2K HD videos captured with a GOPRO4 Hero Black with frame averaging to simulate lower shutter speeds.~\cite{rozumnyi2021defmo} also proposed a synthetic dataset with 3D objects from~\cite{chang2015shapenet} and backgrounds from~\cite{karpathy2014large,kristan2016novel}. The trajectories of objects were uniformly sampled from fixed bounds. Works have also studied VFI for specific domains such as animations~\cite{siyao2021deep}.~\cite{choi2020channel} introduced benchmarks under large motion conditions with 20 240-fps videos sourced from YouTube. Recently,~\cite{sim2021xvfi} introduced a high-resolution high-frame rate benchmark for video interpolation and super-resolution. It includes a total of 4,423 videos recorded with a Phantom Flex4K. 

Most similar to LAVIB, adjacent efforts that compile 4K video datasets~\cite{montgomery1994xiph,niklaus2020softmax,sim2021xvfi,stergiou2022adapool} source videos from media in which professional equipment are used; e.g. movies~\cite{montgomery1994xiph,niklaus2020softmax} or high-resolution video recordings~\cite{sim2021xvfi}. Videos from these datasets are primarily recorded with sensors under optimal shutter speeds and calibrated luminance for capturing specific motion types. In contrast, LAVIB includes videos from various sensors such as hand-held, action, professional, or drone cameras, and screen captures. The videos differ in their dynamic range, levels of post-processing, and compression. LAVIB is intended as a general-purpose dataset and benchmark without being specific to sensor types or settings. Examples of videos are shown in~\Cref{fig:teaser}. 

\looseness-1 LAVIB is compared in~\Cref{tab:datasets} to adjacent video datasets over different statistics. \textbf{Dataset statistics} include the number of videos and total running times.~\textbf{Video statistics} relate to video information such as the resolution and frame rate.~\textbf{Average video metrics} provide metrics on the variance of motions, lighting conditions, and frame sharpness. Definitions of the metrics are detailed in~\S\ref{sec:statistics}. LAVIB has threefold more videos than~\cite{xue2019video} and equally larger total video running time than~\cite{soomro2012ucf101}. The difference in LAVIB video conditions and recording sensors is reflected by the high variance across metrics in~\Cref{tab:datasets}. With the exception of~\cite{sim2021xvfi}, tailored for videos of fast motions with high optical flow magnitude, LAVIB has the highest variance per metric across datasets.

\newcolumntype{d}{>{\columncolor{col1_}}c}
\newcolumntype{e}{>{\columncolor{col1_}}r}

\newcolumntype{f}{>{\columncolor{col2_}}c}
\newcolumntype{g}{>{\columncolor{col2_}}r}

\newcolumntype{h}{>{\columncolor{col3_}}c}
\newcolumntype{i}{>{\columncolor{col3_}}r}

\begin{table}[t]
\centering
\caption{\textbf{Datasets}. Compared to prior efforts, LAVIB provides a large-scale general-purpose dataset of standardized 4K 60 fps videos. It features a significant variance across Average Flow Magnitude (AFM), Average Relevant Luminance (ARL), and Average Laplacian Variances (ALV) in videos.}
\resizebox{\linewidth}{!}{%
\setlength\tabcolsep{2.0pt}
\begin{tabular}{l eee gg iii}
\toprule
 \multirow{2}{*}{Dataset} & \multicolumn{3}{d}{Dataset statistics} & \multicolumn{2}{f}{Video statistics} & \multicolumn{3}{h}{Average video metrics} \\ \cline{2-4} \cline{5-6} \cline{7-9} 
 & Year & Tot. Mins & Tot. Vids & Src Res. & FPS & AFM & ARL & ALV \tstrut \\
 \midrule
 UCF101~\cite{soomro2012ucf101} & 2012 & 1,600 & 13,320 & 240p & 25 & 2.43 $\pm$ 1.85 & 53.37 $\pm$ 13.42 & 53.99 $\pm$ 18.37  \\
 Xiph~\cite{montgomery1994xiph,niklaus2020softmax} & 2020 & 4 & 19 & \textbf{2160p} & 60 & 26.21 $\pm$ 25.19 & 60.64 $\pm$ 10.77 & 95.24 $\pm$ 62.32   \\
 Inter4K~\cite{stergiou2022adapool} & 2021 & 83 & 1,000 & \textbf{2160p} & 60 & 56.38 $\pm$ 14.34 & 56.79 $\pm$ 14.48 & 25.05 $\pm$ 24.05  \\
 X4K1KFPS~\cite{sim2021xvfi} & 2021 & 191 & 4,423 & \textbf{2160p} & \textbf{960} & 266.87 $\pm$ 178.72 & 53.95 $\pm$ 12.07 & 135.67 $\pm$ 78.19  \\
 Vimeo90K~\cite{xue2019video} & 2017 & 356 & 91,701 & 720p & 30 & 49.63 $\pm$ 18.32 & 59.68 $\pm$ 20.89 & 26.26 $\pm$ 29.25   \\
 LAVIB (ours) & 2024 & \textbf{4,660} & \textbf{283,484} & \textbf{2160p} & 60 & 63.10 $\pm$ 58.41 & 38.34 $\pm$ 28.69 & 199.78 $\pm$ 197.79 \\
\bottomrule
\end{tabular}
}
\label{tab:datasets}
\vspace{-1.2em}
\end{table}

\begin{table}[t]
\centering
\caption{\textbf{LAVIB split statistics}. Details per metric for each split.}
\resizebox{\linewidth}{!}{%
\setlength\tabcolsep{8.5pt}
\begin{tabular}{c l l l l l }
\toprule
 \multicolumn{2}{l}{Statistic} & Train & Val & Train+Val & Test \\
 \midrule
 \multirow{2}{*}{~\motion~} & \# Low Flow Mag & 19,605 (10.3\%) & 3,846 (9.3\%) & 23,451 (10.1\%) & 4,898 (9.1\%) \\
 & \# High Flow Mag & 18,976 (10.1\%) & 3,891 (9.4\%) & 22,867 (9.9\%) & 5,482 (10.2\%) \\
 \midrule
 \multirow{2}{*}{~\blur~} & \# Low Lap. Var. & 18,313 (9.6\%) & 3,541 (8.6\%) & 21,854 (9.5\%) & 6,494 (12.1\%) \\
 & \# High Lap. Var. & 17,348 (9.2\%) & 3,871 (9.4\%) & 21,219 (9.2\%) & 7,130 (13.3\%) \\
 \midrule
 \multirow{2}{*}{~\lum~} & \# Low Perc. Lum. & 17,669 (9.3\%) & 3,638 (8.8\%) & 21,307 (9.2\%) & 7,041 (13.1\%) \\
 & \# High Perc. Lum. & 19,297 (10.2\%) & 4,400 (10.7\%) & 23,697 (10.3 \%) & 4,652 (8.6\%) \\
 \midrule
 \multirow{2}{*}{~\contrast~} & \# Low RMS Cont. & 18,794 (10.0\%) & 3,657 (8.8\%) & 22,451 (9.8\%) & 5,897 (11.0\%) \\
 & \# High RMS Cont. & 18,363 (9.7\%) & 4,036 (9.8 \%) & 22,399 (9.7\%) & 5,950 (11.1\%) \\
 \midrule
 \multicolumn{2}{c}{Total} & 188,644 & 41,345 & 229,989 & 53,494 \\
\bottomrule
\end{tabular}
}
\label{tab:splits}
\vspace{-1.2em}
\end{table}

\section{LAVIB statistics}
\label{sec:statistics}

Four statistics are used to obtain segments, create splits, and define challenges. An overview is shown in \Cref{tab:splits} with the number of videos with the highest/lowest metrics reported.

\textbf{Frame-pair motion}. A significant challenge for VFI methods is learning to model the cross-frame motion consistency of videos. Thus, the proposed dataset includes videos of diverse magnitudes; both high camera or object motion, and more static scenes. Motion magnitudes can be quantified with dense optical flow. FlowFormer~\cite{huang2022flowformer} is used on each frame pair resulting in 598 frame pairs per video. The spatial resolution of videos is reduced by $\times0.25$ to fit frames in memory. The Averaged Flow Magnitude (AFM) is defined by spatio-temporally averaging optical flow. AFM variances are reported for all datasets in \Cref{tab:datasets}.     

\textbf{Frame sharpness}. Sourced videos vary by the sensors, lens, codex, and camera profiles used. They can capture different motions, light conditions, and camera focus. All these factors amount to significant variations in the sharpness of videos. Thus, object edges or sensory noise may be highlighted or suppressed. The Laplacian of Gaussians (LoG) is a standardized kernel-based approach for highlighting regions of rapid change in pixel intensities. Given a video $\mathbf{V}$ of dimensions $\mathbb{R}^{D=T \times H \times W}$, with $T$ frames, $H$ height, and $W$ width, it convolves a kernel with size $K$ over each frame. ALV is formulated by applying LoG and averaging:

\begin{gather}
\begin{aligned}
    \text{ALV}(\mathbf{V},\sigma,K) &= \frac{1}{D}\sum_{r \in \mathbb{R}^{D}} \sum_{i=1}^{K}\sum_{j=1}^{K} \underbrace{-1\frac{1}{\pi \sigma^4} (1-\frac{i^2+j^2}{2\sigma^2})e^{-\frac{i^2+j^2}{2\sigma^2}}}_{\text{LoG}(i,j) \text{ kernel}} \; \underset{r-[i,j]}{\mathbf{V}}
\end{aligned}
\end{gather}

As the size of the kernel also factors the estimate, an ensemble of kernel sizes $\mathcal{N}=\{3, 5, 7\}$ is used to calculate the final value $\frac{1}{|\mathcal{N}|}\underset{K\in \mathcal{N}}{\sum}\text{ALV}(\mathbf{V},\sigma,K)$ with $\sigma=1.4$. Overall, in LAVIB 18,313 train, 3,541 val, and 6,494 test videos are at the upper 10\% of the ALV ensemble.

\textbf{Video contrast}. Another characteristic of videos is the contrast between objects and backgrounds in scenes. The human visual system is more sensitive to the contrast between foreground and background~\cite{mazade2022cortical,rahimi2023luminance}, compared to other adjacent measures such as the perceived luminance (brightness), or frame sharpness (blurriness). Computationally, contrast relates to the difference between neighboring raw pixel values. The metric is formulated as the \emph{Average Root Mean Square} (ARMS)~\cite{peli1990contrast} difference between each pixel from each frame of $\mathbf{V}$ and the corresponding pixel in the channel-averaged $\overline{\mathbf{V}}$.  

\begin{equation}
    \text{ARMS}(\mathbf{V}) = \frac{1}{T} \sum_{t\in \mathbb{R}^{T}}\sqrt{\frac{1}{HW}\sum_{s \in \mathbb{R}^{HW}}(\underset{t,s}{\mathbf{V}}-\underset{t,s}{\overline{\mathbf{V}}})}
\end{equation}

LAVIB includes 22,399 videos of high contrast for train and val and, 22,451 videos of low contrast.

\textbf{Luminance conditions}. In addition to the overall video conditions, the perception of light can be affected by the sensor's sensitivity or the camera's processing. In human vision, the perception of luminance is done over three bands of color. To account for the uneven perception of each band, a common standard for quantitatively defining luminosity is the relevant luminance~\cite{poynton2012digital}. In videos, the Average Relative Luminance (ARL) can be computed as the weighted sum for each color channel from video frames based on~\cite{poynton2012digital}, which in turn is averaged over time. The bottom 10th ARL percentile in LAVIB includes 17,669 train, 3,638 val, and 7,041 test videos. Similarly, there are 19,297 train, 4,400 val, and 4,652 test high-luminance videos.

As shown in \Cref{tab:splits}, videos selected for all splits are balanced across metrics. This is done explicitly for the main benchmark and not the OOD challenges.

\begin{figure}[t]
    \centering
    \includegraphics[width=\linewidth]{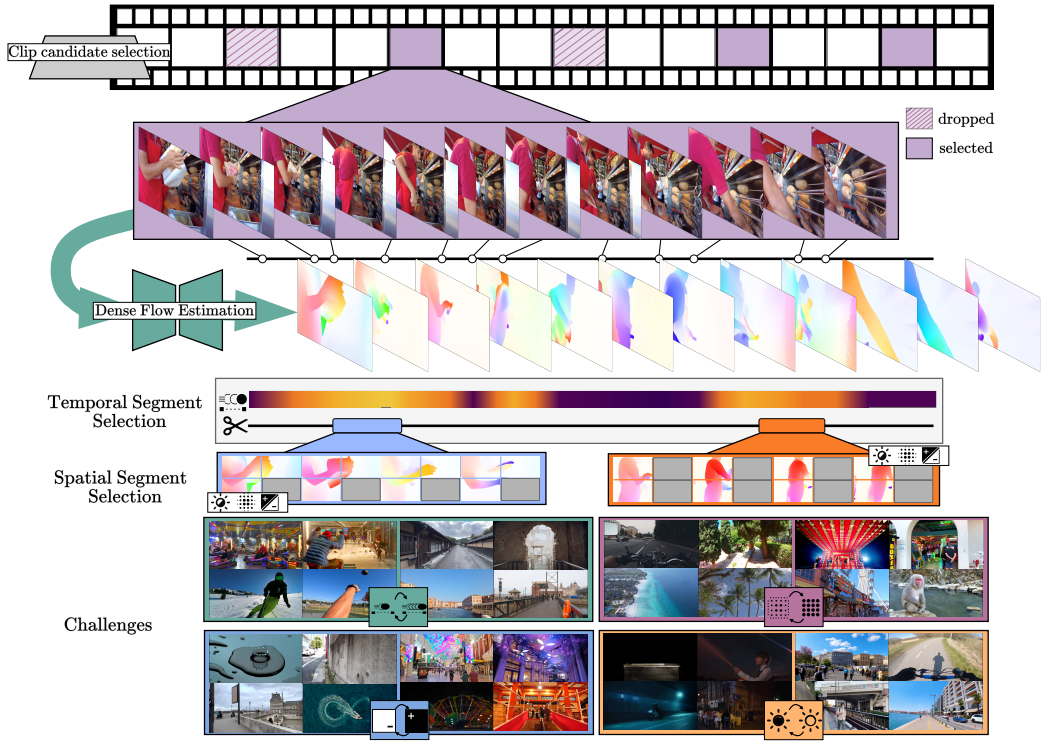}
    \caption{\textbf{LAVIB segment selection and challenges pipeline}. Candidate 10-second clips are sampled from a long video based on their embedding similarity. Dense optical flow is computed with~\cite{huang2022flowformer} and spatially averaged for the AFM metric. The 1-second clips with the top-20\% AFM are selected for the next step. Clips are further partitioned into four tubelets used in the final dataset based on their ARL, ALV, ARMS, and AFM. The metrics are also used for video selection in OOD challenges for a. motion, b. sharpness, c. contrast, d. luminance.} 
    \label{fig:annotation}
    \vspace{-1.2em}
\end{figure}

\section{LAVIB pipeline}
\label{sec:annotation}

The video selection pipeline includes several stages for the collection, extraction, and set assignment. Initially, videos are searched on YouTube by textual prompts designed to return relevant videos with 4K resolutions and 60 frames per second as overviewed in \S\ref{sec:annotation:crawling}. Sourced videos are cropped to 10-second clips standardizing their durations and improving processing speeds in further steps of the collection pipeline. Segments with high motion magnitudes are selected from the clips and are cropped to tubelets by their AVL, ARL, ARMS, and AFM statistics as detailed in \S\ref{sec:annotation:selection}. Dataset splits are balanced between the four statistics, with OOD splits created by assigning videos with the highest average metric at the test or train set. The dataset pipeline is flexible and can be scaled over large numbers of videos, requiring manual input only at a few points.

\subsection{Video web-crawling}
\label{sec:annotation:crawling}

The first stage of the data collection pipeline constructs queries to search and identify videos on YouTube with 4K resolution and 60 fps. The vocabulary of search terms is created from a finite combination of different categories e.g.; locations, activities, weather conditions, and camera types. This aims to diversify results over the defined categories with a level of control (See appendix \ref{app:vocabulary} for a full discussion on vocabulary creation). The vocabulary terms are compiled with three guidelines.

\noindent
\looseness-1 \textbf{Videos should be in the wild}. Retrieved videos should vary by lighting conditions, motions, and scenes. They should also be recorded with different sensors. Sensor types depend strongly on video themes; e.g. action cameras are more common for capturing fast-paced scenes in contrast to DSLR cameras. Conditions are added in the format; \texttt{rainy walk in New York} or \texttt{night drive}. Some text prompts are also designed to include specific equipment such as \texttt{GoPro Hero10} or \texttt{iPhone 13 Pro}. 

\noindent
\textbf{Video content should correspond to raw footage}. The dictionary of general search terms aims to improve control over the video context by retrieving specific video types. Videos with substantial post-production cuts, or transitions, can be less relevant or usable for VFI. The video types that are collected focus primarily on raw footage. 

\textbf{Exclusivity of video categories}. Vocabulary queries should also include diversity in the themes present. This is done by constructing verb hierarchies. A balanced number of queries is constructed for objects/locations that are the focus of the videos.

Each vocabulary search term is combined with `4K' and used as a query on YouTube. The query-related \texttt{URL}s are scraped from the contents in the response's script. Candidate videos are downloaded only if a 4K format with 60 fps is available. This step is needed as YouTube's search prioritizes video elements such as titles, tags, and descriptions over metadata.  

\looseness-1 \textbf{Limitations}. Queries are created from a finite set of search terms. The diversity of locations and activities is manually defined thus, limitations are expected. As noted above, the selected videos are more diverse than current VFI benchmarks however, an increased vocabulary can improve this further.

\subsection{Segment selection and split assignment}
\label{sec:annotation:selection}

In total, 667 hours of footage are collected over the project's 31-month duration. This initial list contained videos of hour-long to minute-long durations. To standardize their durations, 10-second clips are sampled manually over different interval steps. Clips are extracted consecutively for videos less than 5 minutes. For the rest of the videos; 10-second sampling intervals are used for videos with durations between 5-30 minutes, 2-minute intervals for videos between 30 minutes to an hour, and 10-minute intervals for videos longer than an hour. This selection resulted in a total of 34,408 clips. Clips from the same video are bound to include similarities. To account for this and inspired by~\cite{zhang2018unreasonable}, similarities between clips from the same video are measured metrically by their embedding space distance with highly similar clips being dropped. MViTv2-B~\cite{feichtenhofer2022masked} is used to encode clips and to create a similarity matrix based on the L2 distance of the final layer embeddings. Clips with an average (row-wise) L2 distance below the entire matrix's average distance are dropped. This step resulted in the selection of 17,204 clips. 

The final two stages include both temporal and spatial cropping. They are overviewed in \Cref{fig:annotation}.

\textbf{Temporal segment selection}. Segments are compared and selected by their AFM. This selection aims to drop primarily static segments as they are less relevant to VFI tasks with minimal pixel and object tracking requirements. FlowFormer~\cite{huang2022flowformer} is used to calculate AFM over pairs of frames by spatially averaging flows. Each 10-second sequence is temporally augmented to obtain all available 1-second clips. Clips with the highest 20\% magnitudes are selected. This strategy was chosen as it worked well in a small-scale setting when manually examining a set of 1,000 clips.

\textbf{Spatial segment selection}. The selected high-resolution clips cannot directly fit into the memory of most current GPUs. Thus, as commonly addressed in the literature \cite{montgomery1994xiph,niklaus2020softmax,sim2021xvfi,xue2019video} the number of videos is curated with the additional selection of tubelets. Each clip is divided into four tubelets by a $2\times2$ grid. ALV, ARL, ARMS, and AFM are computed for each tubelet.  80\% of the tubelets are retained by selecting from the low/high values per metric in succession, leaving out the middle 20\%. This avoids oversampling from values close to the mean of metrics. Instead, tubelets with more challenging settings are selected.  

\textbf{Assignment to splits}. All train/val/test splits are constructed with a 65-15-20\% split. DUPLEX~\cite{snee1977validation} selection is used for balancing split statistics. Videos with the largest pair-wise distance by their metrics are initially selected. In turn, videos are iteratively assigned to sets given their distance from the previously selected videos. A detailed overview of the algorithm is provided in \S\ref{app:sort}.
Recall that the OOD sets need to be imbalanced across statistics so this is specific to the benchmark splits. 

\textbf{Limitations}. No prior work has tackled video collection based on these metrics, so thresholds for each step are manually defined. This can constrain the final dataset size as the values were selected empirically to maximize diversity.

\section{Benchmarks}
\label{sec:results}

\looseness-1 \textbf{Baselines}. LAVIB contains 188,644 1-second videos for training, 41,345 videos for validation, and 53,494 videos for testing. Benchmark results are reported in \S\ref{sec:results::baseline} across settings. For the baselines, triplets of frames are defined similarly to~\cite{choi2020channel,soomro2012ucf101,xue2019video} for single-frame interpolation with a total of $\sim$5.7M triplets. In the multi-frame interpolation settings in \S\ref{sec:results::ablations}, septuplets are also used resulting in a total of $\sim2.4$M groups of frames. Ablations on varying video resolutions are presented in \S\ref{sec:results::resolution}.  
In \S\ref{sec:results::OOD}, the video metrics are used to create \emph{unbalanced} dataset splits. For each of the four metrics, two challenges are created by sampling videos with either high/low values and assigning them to the train/test sets. Qualitative results for all three models are shown in~\S\ref{sec:results::qualitative}.

\textbf{Model details}. Three VFI methods are benchmarked; RIFE~\cite{huang2022real}, EMA-VFI~\cite{zhang2023extracting}, and FLAVR~\cite{kalluri2023flavr}, which in turn are trained and tested on LAVIB.
The official codebases made publicly available by their respective authors are adjusted and used for LAVIB for all experiments. Adapted training and test code, and models are available at \url{https://github.com/alexandrosstergiou/LAVIB}.

\textbf{Training details}. The training and model settings are imported from the original papers and codebases. The train batch size is set to 64 for all models and the start learning rate is reduced by $\times0.25$ for all models to account for the increased batch size.

\textbf{Evaluation metrics}. Standard image and video quality metrics are used for all tasks and benchmarks. Quantitative results report the Peak Signal-to-Noise Ratio (PSNR), Structural Similarity (SSIM), and Learned Perceptual Image Patch Similarity (LPIPS)~\cite{zhang2018unreasonable}. In multi-frame interpolation, the average value over the interpolated frames is reported.

\begin{table}[t]
\centering
\caption{\textbf{LAVIB val and test results} using~\cite{kalluri2023flavr} as a baseline across training schemes. Evaluation metrics are reported for both val and test sets. Best results per metric are denoted in \textbf{bold}.}
\resizebox{\linewidth}{!}{%
\begin{tabular}{c c c ccc  ccc }
\toprule
 \multirow{2}{*}{\makecell{Pre-train\\ LAVIB}} & \multirow{2}{*}{\makecell{Pre-train\\ Vimeo-90K}} & \multirow{2}{*}{\makecell{Fine-tune \\ Xiph + X4K1KFPS}} & \multicolumn{3}{c}{LAVIB val performance} & \multicolumn{3}{c}{LAVIB test performance} \\
 &  & & PSNR$\uparrow$ & SSIM$\uparrow$ & LLIPS$\downarrow$ & PSNR$\uparrow$ & SSIM$\uparrow$ & LLIPS$\downarrow$ \\
 \midrule
  \rowcolor{LightGrey} & \ding{51} & & 32.86 & 0.968 & 3.152e$^{\!\!-2}$ 
  & 32.10 & 0.963 & 3.947e$^{\!\!-2}$  \\
  & \ding{51} & \ding{51} & 31.36 & 0.952 & 4.620e$^{\!\!-2}$ 
  & 31.78 & 0.948 & 5.154e$^{\!\!-2}$  \\
  \ding{51} & & & \textbf{33.72} & \textbf{0.981} & \textbf{2.515e}$^{\!\!-2}$ 
  & \textbf{33.44} & \textbf{0.981} & \textbf{2.934e}$^{\!\!-2}$ \\
\bottomrule
\end{tabular}
}
\label{tab:main_results_vfi}
\vspace{-1.2em}
\end{table}

\begin{table}[t]
\centering
\begin{minipage}{.325\linewidth}
\caption{\textbf{Vimeo-90K performance}~\cite{kalluri2023flavr} with dif. train sets.}
\resizebox{\linewidth}{!}{%
\setlength\tabcolsep{1pt}
\begin{tabular}{l lll}
\toprule
 Train set & PSNR$\uparrow$ & SSIM$\uparrow$ & LPIPS$\downarrow$ \\
 \midrule
 \rowcolor{LightGrey} Vimeo-90K & 36.25 & 0.975 & $\;\;$9.280e$^{\!\!\!-3}$$^{\textcolor{BrickRed}{*}}$ \bstrut\\
 LAVIB & \textbf{36.68}$\;\;$ & \textbf{0.983}$\;\;$ & $\;\;$\textbf{4.162e}$^{\!\!\!-3}$ \\
\bottomrule
\end{tabular}
\label{tab:vfi_vimeo}
}
\end{minipage}%
\hfill
\begin{minipage}{.325\linewidth}
\caption{\looseness-1 \textbf{Xiph4K performance}~\cite{kalluri2023flavr}  with dif. train sets.}
\resizebox{\linewidth}{!}{%
\setlength\tabcolsep{1pt}
\begin{tabular}{l lll}
\toprule
 Train set & PSNR$\uparrow$ & SSIM$\uparrow$ & LPIPS$\downarrow$ \\
 \midrule
 \rowcolor{LightGrey} Vimeo-90K & 33.28$^{\textcolor{BrickRed}{*}}$ & 0.892$^{\textcolor{BrickRed}{*}}$ & 0.236e$^{\!\!\!-1}$$^{\textcolor{BrickRed}{*}}$ \bstrut\\
 LAVIB & \textbf{34.51}$\;\;$ & \textbf{0.911}$\;\;$ & \textbf{0.532e}$^{\!\!\!-2}$$\;\;$ \\
\bottomrule
\end{tabular}
\label{tab:vfi_xiph}
}
\end{minipage}%
\hfill
\begin{minipage}{.325\linewidth}
\caption{\textbf{X4K1KFPS performance}~\cite{kalluri2023flavr} with dif. train sets.}
\resizebox{\linewidth}{!}{%
\setlength\tabcolsep{1pt}
\begin{tabular}{l lll}
\toprule
 Train set & PSNR$\uparrow$ & SSIM$\uparrow$ & LPIPS$\downarrow$ \\
 \midrule
 \rowcolor{LightGrey} Vimeo-90K & 31.25$^{\textcolor{BrickRed}{*}}$ & 0.9083$^{\textcolor{BrickRed}{*}}$ & 0.383e$^{\!\!\!-1}$$^{\textcolor{BrickRed}{*}}$ \bstrut \\
 LAVIB & \textbf{32.44}$\;\;$ & \textbf{0.927}$\;\;$ & \textbf{0.894e}$^{\!\!\!-2}$$\;\;$ \\
\bottomrule
\end{tabular}
\label{tab:vfi_x4k1kfps}
}
\end{minipage}
\vspace{-1.2em}
\end{table}

\begin{table}[t!]
\centering
\caption{\textbf{Multi-metric evaluation results on LAVIB test}. Performance is reported for $^\medstar$image-based metrics averaged across frames and $^\blacklozenge$video-based metrics.}
\resizebox{\linewidth}{!}{%
\setlength\tabcolsep{2.5pt}
\begin{tabular}{l cccccc ccc }
\toprule
 Model & $^\medstar$PSNR$\uparrow$ & $^\medstar$SSIM$\uparrow$ & $^\medstar$LPIPS$\downarrow$~\cite{zhang2018unreasonable} & $^\medstar$DISTS$\downarrow$~\cite{ding2020image} & $^\medstar$Watson-DFT$\uparrow$~\cite{czolbe2020loss} & $^\blacklozenge$VSFA$\uparrow$~\cite{li2019quality} & $^\blacklozenge$VFIPS$\uparrow$~\cite{hou2022perceptual} \\
 \midrule
 RIFE & 27.88 & 0.871 & 1.416e$^{\!\!-1}$ & 1.870e$^{\!\!-1}$ & 0.215 & 0.558 & 0.561\\
 EMA-VFI & 33.14 & 0.978 & 3.105e$^{\!\!-2}$ & 5.076e$^{\!\!-2}$ & 0.344 & 0.607 & 0.638\\
 FLAVR & \textbf{33.44} & \textbf{0.981} & \textbf{2.934e}$^{\!\!-2}$ & \textbf{4.430e}$^{\!\!-2}$ & \textbf{0.360} & \textbf{0.626} & \textbf{0.667} \\
\bottomrule
\end{tabular}
}
\label{tab:results_vfi_multiscores}
\vspace{-1.1em}
\end{table}

\subsection{Baseline results}
\label{sec:results::baseline}

\let\thefootnote\relax\footnotetext{$^{\textcolor{BrickRed}{*}}$Inhouse evaluation from author provided model.}

\textbf{Baselines}. \Cref{tab:main_results_vfi} reports SSIM, PSNR, and LPIPS scores on both LAVIB val and test sets across three training settings; pre-training on Vimeo-90K, fine-tuning on a joint set from Xiph~\cite{montgomery1994xiph,niklaus2020softmax} and X4K1KFPS~\cite{sim2021xvfi} of exclusively 4K videos, and pre-training with LAVIB. FLAVR~\cite{kalluri2023flavr} is used as the baseline model due to its fast processing times, strong results, and open-source codebase. Finetuning on Xiph + X4K1KFPS suffers as both datasets are small in size although they are sourced by videos with the same resolution as LAVIB. Pre-training only on Vimeo-90K slightly improves results. Pre-training on LAVIB gives the best performance overall increasing PSNR, and SSIM by +1.08 and +0.015 on average on both sets.

\textbf{Generalization to related small-scale datasets}. VFI benchmarks include multiple datasets~\cite{montgomery1994xiph,niklaus2020softmax,sim2021xvfi,xue2019video}. LAVIB is unique in having the largest number of diverse videos of \textit{both} high resolution and high frame rates. The generalization benefits of using LAVIB as the pre-training dataset are compared to the previously widely-used Vimeo-90K~\cite{xue2019video}. \Cref{tab:vfi_vimeo} shows performance improvements in the test set of Vimeo-90K when the model is trained on LAVIB. Similar score increases are also observed for the Xiph4K and X4K1KFPS test sets in \Cref{tab:vfi_xiph,tab:vfi_x4k1kfps} with +1.23 and +1.19 improvements on the PSNR.  LAVIB's large variance across videos enables learning VFI over different conditions which can benefit performance in smaller domain-specific benchmark datasets.

\textbf{Multi-metric results}. As human judgment of the perceptual quality depends on high-order image structures and context~\cite{markman2005nonintentional,wang2004image}, an ensemble of metrics is reported in~\Cref{tab:results_vfi_multiscores} to provide a complete evaluation of each methods' performance on the LAVIB test set. In addition to standard quality metrics, scores over recently-proposed metrics including DISTS~\cite{ding2020image}, Watson-DFT~\cite{czolbe2020loss}, VSFA~\cite{li2019quality}, and VFIPS~\cite{hou2022perceptual} are also reported. Across statistics, both EMA-VFI and FLAVR perform comparably. A decrease in performance is observed with RIFE as its limited complexity can not adequately address VFI with large variations in settings across videos. Compared to FLAVR, the PSNR and SSIM scores decrease by -5.56 and -1.10 respectively, and the LPIPS loss increases from 0.029 to 0.146.

\newcolumntype{G}{>{\columncolor{LightGrey}}c}

\begin{table}[t]
\caption{\textbf{Multi-frame interpolation scores} over triplets, and septuplets across different numbers of interpolated frames. Increase in video duration due to interpolation is denoted with $\{ \times 2, \times 3, \times 4 \}$.}
\resizebox{\linewidth}{!}{%
\setlength\tabcolsep{2.0pt}
\begin{tabular}{l ccc ccc}
\toprule
 \multirow{2}{*}{Model} & 
 \multicolumn{3}{c}{triplet} & \multicolumn{3}{c}{septuplets} \\
 & $\times 2$ & $\times 3$ & $\times 4$ & $\times 2$ & $\times 3$ & $\times 4$ \\
 & PSNR$\uparrow$/SSIM$\downarrow$ & PSNR$\uparrow$/SSIM$\downarrow$ & PSNR$\uparrow$/SSIM$\downarrow$ & PSNR$\uparrow$/SSIM$\downarrow$ & PSNR$\uparrow$/SSIM$\downarrow$ & PSNR$\uparrow$/SSIM$\downarrow$ \\
 \midrule
 \rowcolor{LightGrey} Baseline & 32.10/0.963 & 31.58/0.952 & 30.42/0.937 & 32.69/0.976 & 32.10/0.972 & 31.95/0.918 \\
 FLAVR & \textbf{33.44/0.981} & \textbf{33.07/0.975} & \textbf{32.86/0.968} & \textbf{33.62/0.985} & \textbf{33.41/0.980} & \textbf{33.28/0.962} \\
\bottomrule
\end{tabular}
\label{tab:results_mfi}
}
\vspace{-1.2em}
\end{table}

\begin{table}[!ht]
\centering
\begin{minipage}{.39\linewidth}
\vspace{-0.5em}
\caption{\looseness-1 \textbf{Results on} $\times2$ \textbf{interpolation} with different target fps. Main results default settings in \textcolor{gray}{gray}.}
\resizebox{\linewidth}{!}{%
\setlength\tabcolsep{2.0pt}
\begin{tabular}{c cc GG}
\toprule
 \multirow{2}{*}{\makecell{Model}} & \multicolumn{2}{c}{$15\text{fps} \rightarrow 30\text{fps}$} & 
 \multicolumn{2}{G}{$30\text{fps} \rightarrow 60\text{fps}$} \\
 &  PSNR$\uparrow$ & SSIM$\uparrow$ & PSNR$\uparrow$ & SSIM$\uparrow$  \\
 \midrule
 FLAVR & 33.21 & 0.978 & 33.44 & 0.981 \\
\bottomrule
\\
\end{tabular}
\label{tab:vfi_x2}
}
\end{minipage}%
\hfill
\begin{minipage}{.6\linewidth}
\caption{\textbf{LAVIB test set scores across frame resolutions} with FLAVR on different training schemes. Main results default settings in \textcolor{gray}{gray}.}
\resizebox{\linewidth}{!}{%
\setlength\tabcolsep{2.8pt}
\begin{tabular}{l ccc GGG}
\toprule
 \multirow{2}{*}{\makecell{Train set}} & \multicolumn{3}{c}{$112 \times 112$} & 
 \multicolumn{3}{G}{$256 \times 256$} \\
 &  PSNR$\uparrow$ & SSIM$\uparrow$ & LLIPS$\downarrow$ & PSNR$\uparrow$ & SSIM$\uparrow$ & LLIPS$\downarrow$ \\
 \midrule
 Video90K & 30.14 & 0.943 & 4.638e$^{-2}$ & 32.10 & 0.963 & 3.947e$^{-2}$ \\
 LAVIB & 32.57 & 0.965 & 3.781e$^{-2}$ & \textbf{33.44} & \textbf{0.981} & \textbf{2.934e$^{-2}$} \\
\bottomrule
\end{tabular}
\label{tab:vfi_schemes}
}
\end{minipage}%
\vspace{-1.2em}
\end{table}

\begin{table}[!ht]
\centering
\caption{\textbf{Frame resolutions ablations}. Best results per metric are denoted in \textbf{bold} and best results per model are \underline{underlined}.}
\resizebox{\linewidth}{!}{%
\setlength\tabcolsep{6pt}
\begin{tabular}{l ccc ccc GGG}
\toprule
 \multicolumn{1}{c}{\multirow{2}{*}{Model}} & \multicolumn{3}{c}{$112 \times 112$} & \multicolumn{3}{c}{$224 \times 224$} &
 \multicolumn{3}{G}{$256 \times 256$} \\
 &  PSNR$\uparrow$ & SSIM$\uparrow$ & LLIPS$\downarrow$ & PSNR$\uparrow$ & SSIM$\uparrow$ & LLIPS$\downarrow$ & PSNR$\uparrow$ & SSIM$\uparrow$ & LLIPS$\downarrow$ \\
 \midrule
  EMA-VFI & 32.26 & 0.954 & 4.130e$^{-2}$ & 33.01 & 0.972 & 3.211e$^{-2}$ & \underline{33.14} & \underline{0.978} & \underline{3.105e$^{-2}$}  \\
  FLAVR & 32.57 & 0.965 & 3.781$e^{-2}$ & 33.28 & 0.973 & 3.086$e^{-2}$ & \underline{\textbf{33.44}} & \underline{\textbf{0.981}} & \underline{\textbf{2.934e}$^{-2}$} \\
\bottomrule
\end{tabular}
}
\label{tab:resolutions_vfi}
\vspace{-1.2em}
\end{table}

\begin{table}[t]
    \centering
    \caption{\textbf{PSNR,SSIM, and LPIPS scores on OOD challenges}. Flow-based challenges are denoted by \motionslow~\raisebox{.2em}{$\rightarrow$}\motionfast~for low train and high test AFM and \motionfast~$\raisebox{.2em}{$\rightarrow$}$\motionslow~for high train to low test. For blur-based \blurlow~$\raisebox{.2em}{$\rightarrow$}$\blurhigh~ denotes low and high and \blurhigh~$\raisebox{.2em}{$\rightarrow$}$\blurlow~ denotes high and low. \contrastlow~$\raisebox{.2em}{$\rightarrow$}$\contrasthigh~and\contrasthigh~$\raisebox{.2em}{$\rightarrow$}$\contrastlow~denote low/high, and high/low ARMS respectively. \lumhigh~$\raisebox{.2em}{$\rightarrow$}$\lumlow~and~\lumlow~$\raisebox{.2em}{$\rightarrow$}$\lumhigh~denote low/high, and high/low ARL.}
    \begin{subtable}{.495\linewidth}
      \centering
        \caption{\textbf{AFM}}
        \vspace{-0.5em}
        \resizebox{\linewidth}{!}{%
        \setlength\tabcolsep{1.2pt}
        \label{tab:ood::afm}
        \begin{tabular}{l ccc ccc}
            \toprule
            \multirow{2}{*}{Model} & \multicolumn{3}{c}{\motionslow~\raisebox{.2em}{$\rightarrow$}\motionfast~} & \multicolumn{3}{c}{\motionfast~\raisebox{.2em}{$\rightarrow$}\motionslow~}\\
            & PSNR$\uparrow$ & SSIM$\uparrow$ & LPIPS$\downarrow$ & PSNR$\uparrow$ & SSIM$\uparrow$ & LPIPS$\downarrow$ \\
            \midrule
            RIFE & 25.34 & 0.832 & 3.816e$^{\!\!-1}$ & 28.75 & 0.926 & 8.709e$^{\!\!-2}$ \\
            EMA-VFI & 30.21 & 0.936 & 6.420e$^{\!\!-2}$ & 34.89 & 0.929 & 1.705e$^{\!\!-2}$  \\
            FLAVR & \textbf{30.67} & \textbf{0.959} & \textbf{5.094e}$^{\!\!-2}$ & \textbf{35.66} & \textbf{0.991} & \textbf{1.342e}$^{\!\!-2}$ \\
            \bottomrule
        \end{tabular}
        }
    \end{subtable}%
    \hfill
    \begin{subtable}{.495\linewidth}
      \centering
        \caption{\textbf{ALV}}
        \vspace{-0.5em}
        \resizebox{\linewidth}{!}{%
        \setlength\tabcolsep{1.2pt}
        \label{tab:ood::alv}
        \begin{tabular}{l ccc ccc}
            \toprule
            \multirow{2}{*}{Model} & \multicolumn{3}{c}{\blurhigh~$\raisebox{.2em}{$\rightarrow$}$\blurlow~} & \multicolumn{3}{c}{\blurlow~$\raisebox{.2em}{$\rightarrow$}$\blurhigh~} \bstrut \bstrut \bstrut \\
            & PSNR$\uparrow$ & SSIM$\uparrow$ & LPIPS$\downarrow$ & PSNR$\uparrow$ & SSIM$\uparrow$ & LPIPS$\downarrow$ \\
            \midrule
            RIFE & 26.52 & 0.873 & 1.823e$^{\!\!-1}$ & 29.31 & 0.906 & 9.644e$^{\!\!-2}$ \\
            EMA-VFI & 31.26 & 0.948 & 2.947e$^{\!\!-2}$ & 34.30 & 0.972 & 2.703e$^{\!\!-2}$ \\
            FLAVR & \textbf{31.78} & \textbf{0.962} & \textbf{2.942e}$^{\!\!-2}$ & \textbf{34.67} & \textbf{0.975} & \textbf{2.627e}$^{\!\!-2}$ \\
            \bottomrule
        \end{tabular}
        }
    \end{subtable} 
    \begin{subtable}{.495\linewidth}
      \centering
        \caption{\textbf{ARMS}}
        \vspace{-0.5em}
        \resizebox{\linewidth}{!}{%
        \setlength\tabcolsep{2.0pt}
        \label{tab:ood::arms}
        \begin{tabular}{l ccc ccc}
            \toprule
            \multirow{2}{*}{Model} & \multicolumn{3}{c}{\contrastlow~$\raisebox{.2em}{$\rightarrow$}$\contrasthigh~} & \multicolumn{3}{c}{\contrasthigh~$\raisebox{.2em}{$\rightarrow$}$\contrastlow~}\\
            & PSNR$\uparrow$ & SSIM$\uparrow$ & LPIPS$\downarrow$ & PSNR$\uparrow$ & SSIM$\uparrow$ & LPIPS$\downarrow$ \\
            \midrule
            RIFE & 26.28 & 0.836 & 1.766e$^{\!\!-1}$ & 25.42 & 0.855 & 2.358e$^{\!\!-1}$ \\
            EMA-VFI & 32.79 & 0.964 & 2.930e$^{\!\!-2}$ & 30.65 & 0.951 & 4.467e$^{\!\!-2}$ \\
            FLAVR & \textbf{33.02} & \textbf{0.982} & \textbf{2.561e}$^{\!\!-2}$ & \textbf{31.11} & \textbf{0.977} & \textbf{3.024e}$^{\!\!-2}$ \\
            \bottomrule
        \end{tabular}
        }
    \end{subtable}%
    \hfill
    \begin{subtable}{.495\linewidth}
      \centering
        \caption{\textbf{ARL}}
        \vspace{-0.5em}
        \resizebox{\linewidth}{!}{%
        \label{tab:ood::arl}
        \setlength\tabcolsep{2.0pt}
        \begin{tabular}{l ccc ccc}
            \toprule
            \multirow{2}{*}{Model} & \multicolumn{3}{c}{\lumlow~$\raisebox{.2em}{$\rightarrow$}$\lumhigh~} & \multicolumn{3}{c}{\lumhigh~$\raisebox{.2em}{$\rightarrow$}$\lumlow~}\\
            & PSNR$\uparrow$ & SSIM$\uparrow$ & LPIPS$\downarrow$ & PSNR$\uparrow$ & SSIM$\uparrow$ & LPIPS$\downarrow$ \\
            \midrule
            RIFE & 26.83 & 0.872 & 1.743e$^{\!\!-1}$ & 26.95 & 0.865 & 1.627e$^{\!\!-1}$ \\
            EMA-VFI & 33.55 & 0.974 & 2.723e$^{\!\!-2}$ & 33.41 & 0.968 & 3.031e$^{\!\!-2}$ \\
            FLAVR & \textbf{33.97} & \textbf{0.980} & \textbf{2.543e}$^{\!\!-2}$ & \textbf{34.20} & \textbf{0.976} & \textbf{2.875e}$^{\!\!-2}$ \\
            \bottomrule
        \end{tabular}
        }
    \end{subtable}
    \vspace{-1.1em}
    \label{tab:ood}
\end{table}

\subsection{Multi-frame interpolation results}
\label{sec:results::ablations}

This section ablates the number of frames interpolated and evaluated over different schemes with; $\times 2$ interpolation being equivalent to interpolating 30fps videos to 60fps, $\times 3$ interpolating 20fps to 60fps, and $\times 4$ interpolating 15fps to 60fps. Triplets and septuplets of frames as input are also ablated. Results are reported in \Cref{tab:results_mfi}. FLAVR trained on Vimeo-90K is used as a baseline in all settings. 

\textbf{Varying number of interpolated frames}. The LAVIB-trained model~\cite{kalluri2023flavr} consistently outperforms the baseline trained on Vimeo-90K across different numbers of interpolated frames. An average -1.19/-0.02 PSNR/SSIM drop is observed across $\{ \times 2, \times 3, \times 4 \}$ interpolations when septuplets of frames are used. This drop is more significant for triplets with -1.75/-0.078 PSNR/SSIM.  

\looseness-1 \textbf{Varying number of input frames}. Two settings are used for defining inputs. In triplets, models input a single proceeding and a single succeeding frame with the interpolation target being the in-between frame. In septuplets, two proceeding and two succeeding frames are used as inputs. Models trained with septuplets demonstrate only moderate PSNR/SSIM performance improvements across interpolation settings. This shows that regardless of the input settings the dataset remains challenging.

\looseness-1 \textbf{Varying frame sampling}. LAVIB's standardized 60fps also enables works to explore VFI over more challenging settings with multiple temporal resolutions. \Cref{tab:vfi_x2} reports performance on 30fps targets created by sampling every 2 frames to form triplets. Results show consistency between densely sampling frames sequentially ($30\text{fps} \rightarrow 60\text{fps}$) and sampling with a step of 2 ($15\text{fps} \rightarrow 30\text{fps}$).

\subsection{Varying frame resolution results}
\label{sec:results::resolution}

An important factor for VFI is the clarity of the objects. Different computational budgets can limit availability in training schemes and memory use. 

\noindent
\textbf{Resolutions across models}. Results on different training set resolutions are reported in \Cref{tab:resolutions_vfi}. As in~\cite{huang2022real,kalluri2023flavr,zhang2023extracting}, $256\times 256$ is the standard resolution used for training all models. A proportional decrease in performance is observed at lower resolutions. However, these reductions remain small with an average $-0.14$/$-0.01$ in PSNR/SSIM when using $224 \times 224$ and $-0.87$/$-0.02$ when using $112 \times 112$. Thus, LAVIB can be a suitable benchmark for evaluating low-compute VFI models.   

\noindent
\textbf{Frame resolutions across training schemes}. \Cref{tab:vfi_schemes} reports performances across varying resolutions with different dataset training sets. Compared to the LAVIB-trained model, performance degrades significantly at lower resolutions with the smaller and less diverse Vimeo-90K. The large and varying LAVIB training set can be an effective alternative for training on lower compute resources in which full-resolution videos do not fit in memory. 

\subsection{OOD Challenges}
\label{sec:results::OOD}

OOD challenges aim to test the generalizability of models to domains different from the ones trained. In low to high challenges, train sets include videos of low AFM, ALV, ARMS, or ARL values and the remaining videos of high-value metrics are used for testing. For high to low challenges, train sets have high AFM, ALV, ARMS, or ARL values and test sets have low values.

\textbf{Low/High AFM}. As shown in \Cref{tab:ood::afm}, existing VFI models cannot effectively interpolate frames when trained on videos with low motion magnitudes. Compared to the benchmark results in \Cref{tab:results_vfi_multiscores} a $-2.64$ and $-0.04$ drop is observed for PSNR and SSIM. The embedding distance to ground truth frames also increases by +9.825e$^{\!\!-2}$. In contrast, when models are trained on high motion magnitudes, VFI is easier for the target domain of primarily low magnitudes. The imbalance in performance shows the sensitivity of current models to the motion magnitudes of the training data. 

\textbf{Low/High ALV}. Sharpness-based comparisons are reported in \Cref{tab:ood::alv}. Testing on low-sharpness settings is more challenging for VFI models as object edges are more difficult to define. However, models trained on low-sharpness videos can interpolate high-sharpness videos with an average +2.93 and +0.023 increase in the PSNR and SSIM scores compared to the low-to-high task. 

\textbf{Low/High ARMS}. Results on contrast-based OOD challenges are presented in \Cref{tab:ood::arms}. The domain gap between these two settings is significant. Training on low contrast shows robustness when the domain shifts to high contrast at testing.  However, the same generalization is not observed for the inverse with models trained on high-contrast videos and tested on low-contrast VFI. Compared to low-to-high ARMS, high-to-low ARMS shows a -1.65 drop in PSNR.

\textbf{Low/High ARL}. \Cref{tab:ood::arl} reports performances over brightness settings.
Overall, models from either setting show comparable performance and generalization robustness to the target domain. Minor performance improvements are shown for the high to low task with high luminance training being more effective in cross-domain generalization. 

\begin{figure}[t]
    \centering
    \includegraphics[width=\linewidth]{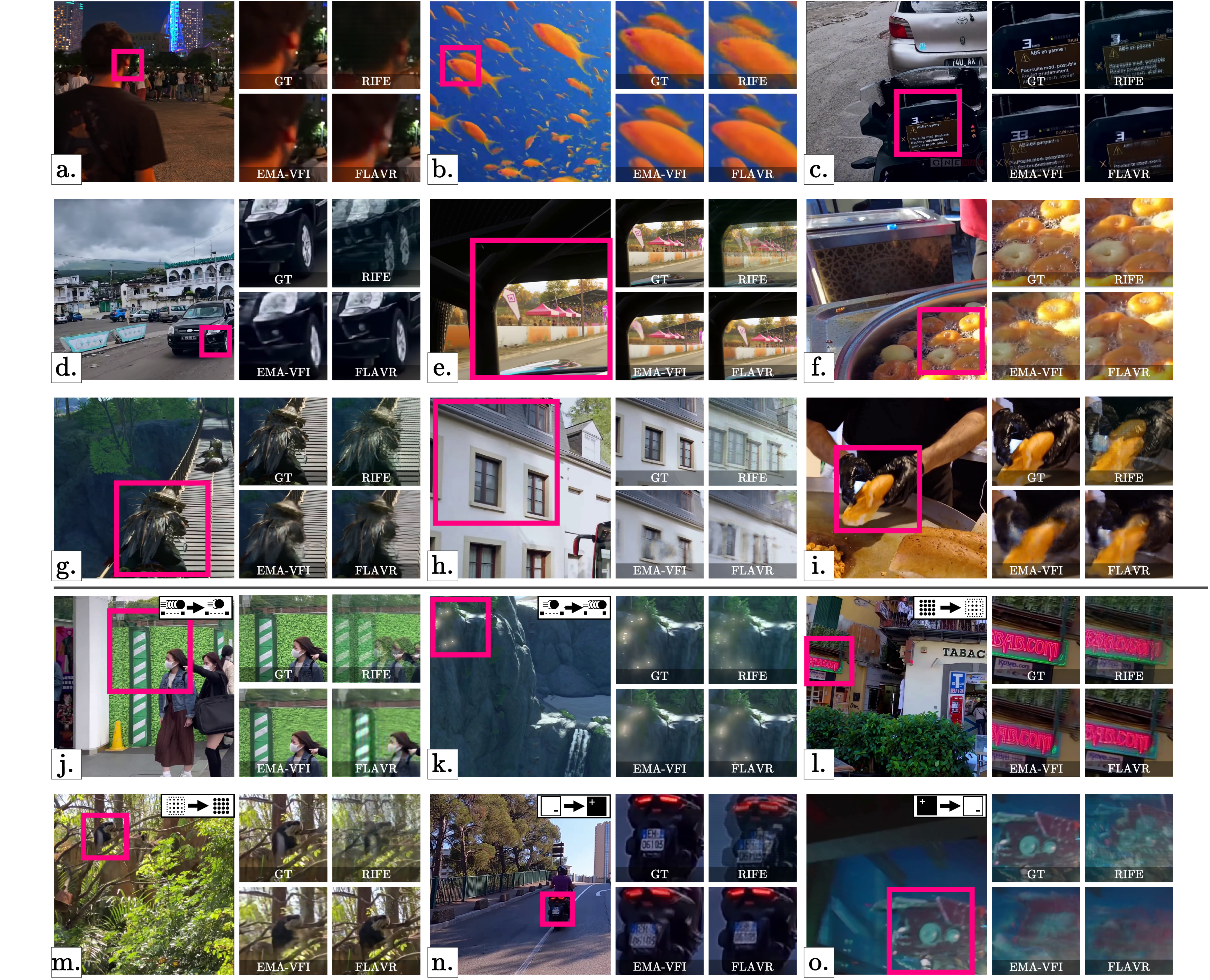}
    \caption{\textbf{Examples from the LAVIB benchmark and OOD test sets} (best viewed digitally). Zoomed regions on the right of each frame show interpolations with RIFE, EMA-VFI, and FLAVR. The top row shows results for videos from the benchmark test split. The bottom two rows are video frames from test splits from OOD challenges. The challenge is denoted at the top right of each ground truth frame. The ground truth is shown as a reference at the top left of the zoomed-in region grid. } 
    \label{fig:qualitative}
    \vspace{-1.2em}
\end{figure}

\subsection{Qualitative results}
\label{sec:results::qualitative}

\looseness-1 \Cref{fig:qualitative} shows interpolated frames from the LAVIB test sets. Frame regions from videos of the LAVIB benchmark interpolated with RIFE, EMA-VFI, and FLAVR are shown in the top three row (\textcolor{red}{a}-\textcolor{red}{i}). Regions shown vary by size and reconstruction error. LAVIB is challenging for current VFI methods as they cannot fully interpolate all parts of objects (\textcolor{red}{b},\textcolor{red}{i}) or fine details (\textcolor{red}{c},\textcolor{red}{f},\textcolor{red}{g}). Objects in scenes affected by high motions are shown to be the most prone to interpolation artifacts as seen with the fine details being missed (\textcolor{red}{d}) and the high cross-frame relative displacement (\textcolor{red}{e}). This also becomes apparent more in high-motion scenes (\textcolor{red}{h}) where large distortions in the scene dynamics can be observed. For OOD challenges models also struggle to correctly interpolate the high contrast between objects and backgrounds (\textcolor{red}{k},\textcolor{red}{l},\textcolor{red}{n}), distinct patterns (\textcolor{red}{j}), and details or objects (\textcolor{red}{m},\textcolor{red}{o}). Further qualitative results are provided in \S\ref{app:qualitative}.

\section{Conclusions and future directions}
\label{sec:conclusions}

\looseness-1 This paper introduces LAVIB, a large-scale general-purpose dataset and benchmark for VFI. LAVIB consists of 283,484 clips collected from 4K videos at 60fps with metrics computed per video specific to motions, sharpness, contrast, and luminance. With the release of the videos and the OOD challenges splits, LAVIB can be used as a robust benchmark and allow the community to investigate VFI under a diverse range of video settings, captured with different equipment, and across various domains.

LAVIB further encourages exploring new avenues for efficiency improvements in future VFI works. 

\textbf{Frame-level quantization}. A number of works have explored video inference acceleration through frame quantization for temporal redundancy reduction   \cite{abati2023resq,sun2021dynamic}. Learning to truncate videos by varying quantization precision is important for the real-world applicability of methods in streams. LAVIB provides a diverse set of high-resolution videos with standardized frame rates that can be used both as a benchmark as well as a pre-training dataset.  

\textbf{Knowledge distillation}. Transferring knowledge about the video structure can enable more efficient models. Several works  \cite{kim2021efficient,ma2022rethinking,purwanto2019extreme} have distilled representations from teacher models trained on high-resolution videos. A natural extension of the proposed dataset would be its use for training high-resolution teacher models and evaluating low-resolution student models.  

\textbf{Salient frame sampling}. The selection of informative frames has been another domain of interest for real-time video processing \cite{wang2022adafocusv3,wu2019adaframe,xia2022nsnet}. The standardized framerate of LAVIB can provide a robust benchmark for testing sampling approaches over different granularities. 

\noindent
Based on these adjacent video tasks, LAVIB can be imported and adapted as a general-purpose dataset and benchmark.

{
\small
\bibliographystyle{ieee_fullname}
\bibliography{refs}

\begin{thebibliography}{10}\itemsep=-1pt

\bibitem{abati2023resq}
Davide Abati, Haitam Ben~Yahia, Markus Nagel, and Amirhossein Habibian.
\newblock Resq: Residual quantization for video perception.
\newblock In {\em ICCV}, pages 17119--17129, 2023.

\bibitem{baker2011database}
Simon Baker, Daniel Scharstein, James~P Lewis, Stefan Roth, Michael~J Black, and Richard Szeliski.
\newblock A database and evaluation methodology for optical flow.
\newblock {\em IJCV}, pages 1--31, 2011.

\bibitem{bao2019depth}
Wenbo Bao, Wei-Sheng Lai, Chao Ma, Xiaoyun Zhang, Zhiyong Gao, and Ming-Hsuan Yang.
\newblock Depth-aware video frame interpolation.
\newblock In {\em CVPR}, pages 3703--3712, 2019.

\bibitem{bovik2017robust}
AC Bovik, R Soundararajan, and Christos Bampis.
\newblock On the robust performance of the st-rred video quality predictor, 2017.

\bibitem{chan2022basicvsr}
Kelvin~CK Chan, Shangchen Zhou, Xiangyu Xu, and Chen~Change Loy.
\newblock Basicvsr++: Improving video super-resolution with enhanced propagation and alignment.
\newblock In {\em CVPR}, pages 5972--5981, 2022.

\bibitem{chang2015shapenet}
Angel~X Chang, Thomas Funkhouser, Leonidas Guibas, Pat Hanrahan, Qixing Huang, Zimo Li, Silvio Savarese, Manolis Savva, Shuran Song, Hao Su, et~al.
\newblock Shapenet: An information-rich 3d model repository.
\newblock {\em arXiv preprint arXiv:1512.03012}, 2015.

\bibitem{choi2020channel}
Myungsub Choi, Heewon Kim, Bohyung Han, Ning Xu, and Kyoung~Mu Lee.
\newblock Channel attention is all you need for video frame interpolation.
\newblock In {\em AAAI}, pages 10663--10671, 2020.

\bibitem{czolbe2020loss}
Steffen Czolbe, Oswin Krause, Ingemar Cox, and Christian Igel.
\newblock A loss function for generative neural networks based on watson’s perceptual model.
\newblock {\em NeurIPS}, pages 2051--2061, 2020.

\bibitem{ding2020image}
Keyan Ding, Kede Ma, Shiqi Wang, and Eero~P Simoncelli.
\newblock Image quality assessment: Unifying structure and texture similarity.
\newblock {\em TPAMI}, pages 2567--2581, 2020.

\bibitem{ding2021cdfi}
Tianyu Ding, Luming Liang, Zhihui Zhu, and Ilya Zharkov.
\newblock Cdfi: Compression-driven network design for frame interpolation.
\newblock In {\em CVPR}, pages 8001--8011, 2021.

\bibitem{feichtenhofer2022masked}
Christoph Feichtenhofer, Yanghao Li, Kaiming He, et~al.
\newblock Masked autoencoders as spatiotemporal learners.
\newblock {\em NeurIPS}, pages 35946--35958, 2022.

\bibitem{hou2022perceptual}
Qiqi Hou, Abhijay Ghildyal, and Feng Liu.
\newblock A perceptual quality metric for video frame interpolation.
\newblock In {\em ECCV}, pages 234--253, 2022.

\bibitem{hu2022you}
Mengshun Hu, Kui Jiang, Zhixiang Nie, and Zheng Wang.
\newblock You only align once: Bidirectional interaction for spatial-temporal video super-resolution.
\newblock In {\em ACM-MM}, pages 847--855, 2022.

\bibitem{huang2022neural}
Cong Huang, Jiahao Li, Bin Li, Dong Liu, and Yan Lu.
\newblock Neural compression-based feature learning for video restoration.
\newblock In {\em CVPR}, pages 5872--5881, 2022.

\bibitem{huang2022flowformer}
Zhaoyang Huang, Xiaoyu Shi, Chao Zhang, Qiang Wang, Ka~Chun Cheung, Hongwei Qin, Jifeng Dai, and Hongsheng Li.
\newblock Flowformer: A transformer architecture for optical flow.
\newblock In {\em ECCV}, pages 668--685, 2022.

\bibitem{huang2022real}
Zhewei Huang, Tianyuan Zhang, Wen Heng, Boxin Shi, and Shuchang Zhou.
\newblock Real-time intermediate flow estimation for video frame interpolation.
\newblock In {\em ECCV}, pages 624--642, 2022.

\bibitem{isobe2022look}
Takashi Isobe, Xu Jia, Xin Tao, Changlin Li, Ruihuang Li, Yongjie Shi, Jing Mu, Huchuan Lu, and Yu-Wing Tai.
\newblock Look back and forth: Video super-resolution with explicit temporal difference modeling.
\newblock In {\em CVPR}, pages 17411--17420, 2022.

\bibitem{ji2020real}
Xiaozhong Ji, Yun Cao, Ying Tai, Chengjie Wang, Jilin Li, and Feiyue Huang.
\newblock Real-world super-resolution via kernel estimation and noise injection.
\newblock In {\em CVPRw}, pages 466--467, 2020.

\bibitem{jiang2018super}
Huaizu Jiang, Deqing Sun, Varun Jampani, Ming-Hsuan Yang, Erik Learned-Miller, and Jan Kautz.
\newblock Super slomo: High quality estimation of multiple intermediate frames for video interpolation.
\newblock In {\em CVPR}, pages 9000--9008, 2018.

\bibitem{jin2019learning}
Meiguang Jin, Zhe Hu, and Paolo Favaro.
\newblock Learning to extract flawless slow motion from blurry videos.
\newblock In {\em CVPR}, pages 8112--8121, 2019.

\bibitem{jin2023unified}
Xin Jin, Longhai Wu, Jie Chen, Youxin Chen, Jayoon Koo, and Cheul-hee Hahm.
\newblock A unified pyramid recurrent network for video frame interpolation.
\newblock In {\em CVPR}, pages 1578--1587, 2023.

\bibitem{jo2018deep}
Younghyun Jo, Seoung~Wug Oh, Jaeyeon Kang, and Seon~Joo Kim.
\newblock Deep video super-resolution network using dynamic upsampling filters without explicit motion compensation.
\newblock In {\em CVPR}, pages 3224--3232, 2018.

\bibitem{kalluri2023flavr}
Tarun Kalluri, Deepak Pathak, Manmohan Chandraker, and Du Tran.
\newblock Flavr: Flow-agnostic video representations for fast frame interpolation.
\newblock In {\em WACV}, pages 2071--2082, 2023.

\bibitem{karpathy2014large}
Andrej Karpathy, George Toderici, Sanketh Shetty, Thomas Leung, Rahul Sukthankar, and Li Fei-Fei.
\newblock Large-scale video classification with convolutional neural networks.
\newblock In {\em CVPR}, pages 1725--1732, 2014.

\bibitem{kim2021efficient}
Hanul Kim, Mihir Jain, Jun-Tae Lee, Sungrack Yun, and Fatih Porikli.
\newblock Efficient action recognition via dynamic knowledge propagation.
\newblock In {\em ICCV}, pages 13719--13728, 2021.

\bibitem{kim2023event}
Taewoo Kim, Yujeong Chae, Hyun-Kurl Jang, and Kuk-Jin Yoon.
\newblock Event-based video frame interpolation with cross-modal asymmetric bidirectional motion fields.
\newblock In {\em CVPR}, pages 18032--18042, 2023.

\bibitem{kristan2016novel}
Matej Kristan, Jiri Matas, Ale{\v{s}} Leonardis, Tom{\'a}{\v{s}} Voj{\'\i}{\v{r}}, Roman Pflugfelder, Gustavo Fernandez, Georg Nebehay, Fatih Porikli, and Luka {\v{C}}ehovin.
\newblock A novel performance evaluation methodology for single-target trackers.
\newblock {\em TPAMI}, pages 2137--2155, 2016.

\bibitem{kupyn2019deblurgan}
Orest Kupyn, Tetiana Martyniuk, Junru Wu, and Zhangyang Wang.
\newblock Deblurgan-v2: Deblurring (orders-of-magnitude) faster and better.
\newblock In {\em ICCV}, pages 8878--8887, 2019.

\bibitem{li2019quality}
Dingquan Li, Tingting Jiang, and Ming Jiang.
\newblock Quality assessment of in-the-wild videos.
\newblock In {\em ACM-MM}, pages 2351--2359, 2019.

\bibitem{li2020video}
Haopeng Li, Yuan Yuan, and Qi Wang.
\newblock Video frame interpolation via residue refinement.
\newblock In {\em ICASSP}, pages 2613--2617, 2020.

\bibitem{li2021comisr}
Yinxiao Li, Pengchong Jin, Feng Yang, Ce Liu, Ming-Hsuan Yang, and Peyman Milanfar.
\newblock Comisr: Compression-informed video super-resolution.
\newblock In {\em ICCV}, pages 2543--2552, 2021.

\bibitem{liang2024vrt}
Jingyun Liang, Jiezhang Cao, Yuchen Fan, Kai Zhang, Rakesh Ranjan, Yawei Li, Radu Timofte, and Luc Van~Gool.
\newblock Vrt: A video restoration transformer.
\newblock {\em T-IP}, pages 2171--2182, 2024.

\bibitem{liang2022recurrent}
Jingyun Liang, Yuchen Fan, Xiaoyu Xiang, Rakesh Ranjan, Eddy Ilg, Simon Green, Jiezhang Cao, Kai Zhang, Radu Timofte, and Luc~V Gool.
\newblock Recurrent video restoration transformer with guided deformable attention.
\newblock {\em NeurIPS}, pages 378--393, 2022.

\bibitem{liu2013bayesian}
Ce Liu and Deqing Sun.
\newblock On bayesian adaptive video super resolution.
\newblock {\em TPAMI}, pages 346--360, 2013.

\bibitem{ma2022rethinking}
Chuofan Ma, Qiushan Guo, Yi Jiang, Ping Luo, Zehuan Yuan, and Xiaojuan Qi.
\newblock Rethinking resolution in the context of efficient video recognition.
\newblock {\em NeurIPS}, pages 37865--37877, 2022.

\bibitem{markman2005nonintentional}
Arthur~B Markman and Dedre Gentner.
\newblock Nonintentional similarity processing.
\newblock {\em The new unconscious}, pages 107--137, 2005.

\bibitem{mazade2022cortical}
Reece Mazade, Jianzhong Jin, Hamed Rahimi-Nasrabadi, Sohrab Najafian, Carmen Pons, and Jose-Manuel Alonso.
\newblock Cortical mechanisms of visual brightness.
\newblock {\em Cell reports}, pages 111438--111438, 2022.

\bibitem{montgomery1994xiph}
Christopher Montgomery and H Lars.
\newblock Xiph. org video test media (derf’s collection).
\newblock {\em Online, https://media. xiph. org/video/derf}, 6, 1994.

\bibitem{nah2019ntire}
Seungjun Nah, Sungyong Baik, Seokil Hong, Gyeongsik Moon, Sanghyun Son, Radu Timofte, and Kyoung Mu~Lee.
\newblock Ntire 2019 challenge on video deblurring and super-resolution: Dataset and study.
\newblock In {\em CVPRw}, pages 1996--2005, 2019.

\bibitem{nah2017deep}
Seungjun Nah, Tae Hyun~Kim, and Kyoung Mu~Lee.
\newblock Deep multi-scale convolutional neural network for dynamic scene deblurring.
\newblock In {\em CVPR}, pages 3883--3891, 2017.

\bibitem{niklaus2020softmax}
Simon Niklaus and Feng Liu.
\newblock Softmax splatting for video frame interpolation.
\newblock In {\em CVPR}, pages 5437--5446, 2020.

\bibitem{niklaus2017video}
Simon Niklaus, Long Mai, and Feng Liu.
\newblock Video frame interpolation via adaptive separable convolution.
\newblock In {\em ICCV}, pages 261--270, 2017.

\bibitem{paliwal2023implicit}
Avinash Paliwal, Andrii Tsarov, and Nima~Khademi Kalantari.
\newblock Implicit view-time interpolation of stereo videos using multi-plane disparities and non-uniform coordinates.
\newblock In {\em CVPR}, pages 888--898, 2023.

\bibitem{park2021asymmetric}
Junheum Park, Chul Lee, and Chang-Su Kim.
\newblock Asymmetric bilateral motion estimation for video frame interpolation.
\newblock In {\em ICCV}, pages 14539--14548, 2021.

\bibitem{peli1990contrast}
Eli Peli.
\newblock Contrast in complex images.
\newblock {\em JOSA A}, pages 2032--2040, 1990.

\bibitem{perazzi2016benchmark}
Federico Perazzi, Jordi Pont-Tuset, Brian McWilliams, Luc Van~Gool, Markus Gross, and Alexander Sorkine-Hornung.
\newblock A benchmark dataset and evaluation methodology for video object segmentation.
\newblock In {\em CVPR}, pages 724--732, 2016.

\bibitem{poynton2012digital}
Charles Poynton.
\newblock {\em Digital video and HD: Algorithms and Interfaces}.
\newblock Elsevier, 2012.

\bibitem{purwanto2019extreme}
Didik Purwanto, Rizard Renanda Adhi~Pramono, Yie-Tarng Chen, and Wen-Hsien Fang.
\newblock Extreme low resolution action recognition with spatial-temporal multi-head self-attention and knowledge distillation.
\newblock In {\em ICCVw}, pages 0--0, 2019.

\bibitem{rahimi2023luminance}
Hamed Rahimi-Nasrabadi, Veronica Moore-Stoll, Jia Tan, Stephen Dellostritto, JianZhong Jin, Mitchell~W Dul, and Jose-Manuel Alonso.
\newblock Luminance contrast shifts dominance balance between on and off pathways in human vision.
\newblock {\em Journal of Neuroscience}, pages 993--1007, 2023.

\bibitem{reda2022film}
Fitsum Reda, Janne Kontkanen, Eric Tabellion, Deqing Sun, Caroline Pantofaru, and Brian Curless.
\newblock Film: Frame interpolation for large motion.
\newblock In {\em ECCV}, pages 250--266, 2022.

\bibitem{rozumnyi2021defmo}
Denys Rozumnyi, Martin~R Oswald, Vittorio Ferrari, Jiri Matas, and Marc Pollefeys.
\newblock Defmo: Deblurring and shape recovery of fast moving objects.
\newblock In {\em CVPR}, pages 3456--3465, 2021.

\bibitem{shangguan2022learning}
Wentao Shangguan, Yu Sun, Weijie Gan, and Ulugbek~S Kamilov.
\newblock Learning cross-video neural representations for high-quality frame interpolation.
\newblock In {\em ECCV}, pages 511--528, 2022.

\bibitem{sheth2021unsupervised}
Dev~Yashpal Sheth, Sreyas Mohan, Joshua~L Vincent, Ramon Manzorro, Peter~A Crozier, Mitesh~M Khapra, Eero~P Simoncelli, and Carlos Fernandez-Granda.
\newblock Unsupervised deep video denoising.
\newblock In {\em ICCV}, pages 1759--1768, 2021.

\bibitem{shi2022rethinking}
Shuwei Shi, Jinjin Gu, Liangbin Xie, Xintao Wang, Yujiu Yang, and Chao Dong.
\newblock Rethinking alignment in video super-resolution transformers.
\newblock {\em NeurIPS}, pages 36081--36093, 2022.

\bibitem{sim2021xvfi}
Hyeonjun Sim, Jihyong Oh, and Munchurl Kim.
\newblock Xvfi: extreme video frame interpolation.
\newblock In {\em ICCV}, pages 14489--14498, 2021.

\bibitem{siyao2021deep}
Li Siyao, Shiyu Zhao, Weijiang Yu, Wenxiu Sun, Dimitris Metaxas, Chen~Change Loy, and Ziwei Liu.
\newblock Deep animation video interpolation in the wild.
\newblock In {\em CVPR}, pages 6587--6595, 2021.

\bibitem{snee1977validation}
Ronald~D Snee.
\newblock Validation of regression models: methods and examples.
\newblock {\em Technometrics}, pages 415--428, 1977.

\bibitem{song2022tempformer}
Mingyang Song, Yang Zhang, and Tun{\c{c}}~O Ayd{\i}n.
\newblock Tempformer: Temporally consistent transformer for video denoising.
\newblock In {\em ECCV}, pages 481--496, 2022.

\bibitem{soomro2012ucf101}
Khurram Soomro, Amir~Roshan Zamir, and Mubarak Shah.
\newblock Ucf101: A dataset of 101 human actions classes from videos in the wild.
\newblock {\em arXiv preprint arXiv:1212.0402}, 2012.

\bibitem{stergiou2022adapool}
Alexandros Stergiou and Ronald Poppe.
\newblock Adapool: Exponential adaptive pooling for information-retaining downsampling.
\newblock {\em T-IP}, pages 251--266, 2022.

\bibitem{su2017deep}
Shuochen Su, Mauricio Delbracio, Jue Wang, Guillermo Sapiro, Wolfgang Heidrich, and Oliver Wang.
\newblock Deep video deblurring for hand-held cameras.
\newblock In {\em CVPR}, pages 1279--1288, 2017.

\bibitem{sun2021dynamic}
Ximeng Sun, Rameswar Panda, Chun-Fu~Richard Chen, Aude Oliva, Rogerio Feris, and Kate Saenko.
\newblock Dynamic network quantization for efficient video inference.
\newblock In {\em ICCV}, pages 7375--7385, 2021.

\bibitem{tao2017detail}
Xin Tao, Hongyun Gao, Renjie Liao, Jue Wang, and Jiaya Jia.
\newblock Detail-revealing deep video super-resolution.
\newblock In {\em ICCV}, pages 4472--4480, 2017.

\bibitem{tassano2019dvdnet}
Matias Tassano, Julie Delon, and Thomas Veit.
\newblock Dvdnet: A fast network for deep video denoising.
\newblock In {\em ICIP}, pages 1805--1809, 2019.

\bibitem{wang2020deep}
Longguang Wang, Yulan Guo, Li Liu, Zaiping Lin, Xinpu Deng, and Wei An.
\newblock Deep video super-resolution using hr optical flow estimation.
\newblock {\em T-IP}, pages 4323--4336, 2020.

\bibitem{wang2022adafocusv3}
Yulin Wang, Yang Yue, Xinhong Xu, Ali Hassani, Victor Kulikov, Nikita Orlov, Shiji Song, Humphrey Shi, and Gao Huang.
\newblock Adafocusv3: On unified spatial-temporal dynamic video recognition.
\newblock In {\em ECCV}, pages 226--243, 2022.

\bibitem{wang2004image}
Zhou Wang, Alan~C Bovik, Hamid~R Sheikh, and Eero~P Simoncelli.
\newblock Image quality assessment: from error visibility to structural similarity.
\newblock {\em T-IP}, pages 600--612, 2004.

\bibitem{wang2018multi}
Zhongyuan Wang, Peng Yi, Kui Jiang, Junjun Jiang, Zhen Han, Tao Lu, and Jiayi Ma.
\newblock Multi-memory convolutional neural network for video super-resolution.
\newblock {\em T-IP}, pages 2530--2544, 2018.

\bibitem{wu2019adaframe}
Zuxuan Wu, Caiming Xiong, Chih-Yao Ma, Richard Socher, and Larry~S Davis.
\newblock Adaframe: Adaptive frame selection for fast video recognition.
\newblock In {\em CVPR}, pages 1278--1287, 2019.

\bibitem{xia2022nsnet}
Boyang Xia, Wenhao Wu, Haoran Wang, Rui Su, Dongliang He, Haosen Yang, Xiaoran Fan, and Wanli Ouyang.
\newblock Nsnet: Non-saliency suppression sampler for efficient video recognition.
\newblock In {\em ECCV}, pages 705--723, 2022.

\bibitem{xu2021moving}
Dan Xu, Andrea Vedaldi, and Joao~F Henriques.
\newblock Moving slam: Fully unsupervised deep learning in non-rigid scenes.
\newblock In {\em IROS}, pages 4611--4617, 2021.

\bibitem{xue2019video}
Tianfan Xue, Baian Chen, Jiajun Wu, Donglai Wei, and William~T Freeman.
\newblock Video enhancement with task-oriented flow.
\newblock {\em IJCV}, pages 1106--1125, 2019.

\bibitem{yi2019progressive}
Peng Yi, Zhongyuan Wang, Kui Jiang, Junjun Jiang, and Jiayi Ma.
\newblock Progressive fusion video super-resolution network via exploiting non-local spatio-temporal correlations.
\newblock In {\em ICCV}, pages 3106--3115, 2019.

\bibitem{yoo2023video}
Jun-Sang Yoo, Hongjae Lee, and Seung-Won Jung.
\newblock Video object segmentation-aware video frame interpolation.
\newblock In {\em ICCV}, pages 12322--12333, 2023.

\bibitem{zhang2023extracting}
Guozhen Zhang, Yuhan Zhu, Haonan Wang, Youxin Chen, Gangshan Wu, and Limin Wang.
\newblock Extracting motion and appearance via inter-frame attention for efficient video frame interpolation.
\newblock In {\em CVPR}, pages 5682--5692, 2023.

\bibitem{zhang2018unreasonable}
Richard Zhang, Phillip Isola, Alexei~A Efros, Eli Shechtman, and Oliver Wang.
\newblock The unreasonable effectiveness of deep features as a perceptual metric.
\newblock In {\em CVPR}, pages 586--595, 2018.

\end{thebibliography}
}

\newpage

\setcounter{section}{0}
\setcounter{equation}{0}
\setcounter{figure}{0}
\setcounter{table}{0}
\Appendix

{\Large LAVIB: A Large-scale Video Interpolation Benchmark -- Appendix}

\begin{table}[ht]

\centering
\caption{\textbf{Vocabulary of search terms}. Terms are grouped to five main types including \textcolor{col1}{location}, \textcolor{col2}{activities}, \textcolor{col3}{weather}, \textcolor{col4}{misc}, and \textcolor{col5}{camera types}. Search queries are the combination of multiple terms with an additional `4K'.}
\resizebox{\linewidth}{!}{%
\begin{tabular}{ccc cc cccc }
\toprule
 \multicolumn{3}{c}{\textcolor{col1}{Location}} & \multicolumn{2}{c}{\makecell{\textcolor{col2}{Activities}}} & \multirow{2}{*}{\textcolor{col3}{Weather}} & \multirow{2}{*}{\textcolor{col4}{Misc}} & \multirow{2}{*}{\textcolor{col5}{Camera types}} \\
 city & region & country & sports & actions \\
 \midrule
 \makecell{[\texttt{Amsterdam},\\ \texttt{Athens},\\ \texttt{Boston},\\ \texttt{Buenos Aires},\\ \texttt{Doha},\\ \texttt{Dubai},\\ \texttt{Istanbul},\\ \texttt{Lagos},\\ \texttt{Las Vegas},\\ \texttt{London},\\ \texttt{Los Angeles},\\ \texttt{Manchester},\\ \texttt{Mexico City},\\ \texttt{Miami},\\ \texttt{Montreal},\\ \texttt{New York},\\ \texttt{Paris},\\ \texttt{Perth},\\ \texttt{Porto},\\ \texttt{Rio de Janeiro},\\ \texttt{Seoul},\\ \texttt{Shanghai},\\ \texttt{Singapore},\\ \texttt{Tokyo},\\ \texttt{Venice},\\ \texttt{Vienna}]} & 
 
 \makecell{[\texttt{Atlantic},\\ \texttt{California},\\ \texttt{Caribbean},\\ \texttt{England},\\ \texttt{Indian Ocean},\\ \texttt{Scandinavia},\\ \texttt{Sedona},\\ \texttt{Sicily},\\ \texttt{South East}]} &
 
 \makecell{[\texttt{Australia},\\ \texttt{Brazil},\\ \texttt{Bulgaria},\\ \texttt{Cambodia},\\ \texttt{Canada},\\ \texttt{China},\\ \texttt{Costa Rica},\\ \texttt{France},\\ \texttt{Germany},\\ \texttt{Iceland},\\ \texttt{India},\\ \texttt{Japan},\\ \texttt{Morocco},\\ \texttt{Mexico},\\ \texttt{Mongolia},\\ \texttt{Namibia},\\ \texttt{New Zealand},\\ \texttt{Nigeria},\\ \texttt{Russia},\\ \texttt{South Africa},\\ \texttt{Spain},\\ \texttt{Thailand}]} &
 
 \makecell{[\texttt{climbing},\\ \texttt{playing football},\\ \texttt{rafting},\\ \texttt{skating},\\ \texttt{skiing},\\ \texttt{snorkeling},\\ \texttt{snowboarding},\\ \texttt{tennis training}]} & 
 
 \makecell{[\texttt{bike ride},\\ \texttt{car ride},\\ \texttt{dancing},\\
 \texttt{exploration},\\ \texttt{walking}]} & 
 
 \makecell{[\texttt{cloudy},\\ \texttt{overcast},\\ \texttt{rainy},\\ \texttt{snowing},\\ \texttt{sunny}] } & 
 
 \makecell{[\texttt{Dolby Vision},\\ \texttt{PS5},\\ \texttt{animals},\\ \texttt{birds},\\ \texttt{flowers},\\ \texttt{forest},\\ \texttt{insects},\\ \texttt{marinelife},\\ \texttt{metro},\\ \texttt{mountains},\\ \texttt{ocean view},\\ \texttt{park},\\ \texttt{shoreline},\\ \texttt{car},\\ \texttt{underwater},\\ \texttt{wildlife},\\ \texttt{windmills}]} & 
 
 \makecell{[\texttt{Blackmagic PCC 4K},\\ \texttt{Canon 5D Mark III},\\ \texttt{Canon EOS C200B},\\ \texttt{Canon EOS R6},\\ \texttt{DJI Inspire2},\\ \texttt{DJI OM4},\\ \texttt{DJI Osmo Pocket},\\ \texttt{GOPRO HERO10 Black},\\ \texttt{GOPRO HERO8 Black},\\ \texttt{GOPRO HERO9},\\ \texttt{GOPRO Max 360},\\ \texttt{Note 10 plus},\\ \texttt{RED RAVEN 4.5K},\\ \texttt{Samsung Galaxy},\\ \texttt{Sony A6700},\\ \texttt{Sony A7C},\\ \texttt{Yi 4K+},\\ \texttt{iPhone 12 Pro},\\ \texttt{iPhone 13 Pro}]}\\
 \bottomrule
 
\end{tabular}
}
\label{tab:results_vfi}
\vspace{-1.2em}
\end{table}

\section{Vocabulary}
\label{app:vocabulary}

Three core components are used for creating search terms from the vocabulary; locations, activities, or specific objects/settings relevant to videos. Locations and activities include two levels of hierarchies. The structure of search terms changes based on the selected sub-group. 

\subsection{Location}

\noindent
\textbf{Motivation}. Natural scenes were found to have a large number of 4K footage from diverse camera types with minimal edits. Using an exhaustive list of locations is not feasible given the search space.

\noindent
\textbf{Remedy used}. Instead, a list of locations was manually created based on the number of returned videos per location. Oversaturation of similar video locations; e.g. same country, was also manually adjusted for the selected terms.

\noindent
\textbf{About}. The \texttt{city} subgroup is combined with a specific set of actions \{\texttt{bike ride}, \texttt{car ride}, \texttt{exploration}, \texttt{walking}\}. Weather conditions are added randomly to 1/3 of the search terms and camera types are added in 1/10, e.g.; `\textcolor{col1}{Amsterdam} \textcolor{col2}{bike ride} \textcolor{col3}{rainy} \textcolor{col5}{GOPRO HERO10 Black} 4K'. It was seen that camera-type prompts can return results more relevant to the camera (e.g. reviews) and less relevant to the rest of the term searched. Thus, the probability of including camera types is kept low. For the \texttt{region} and \texttt{country} subgroups, prompts only include keywords such as `best of' or `scenic' as actions are less relevant when the locations are broad.

\noindent
\textbf{Limitations} The manually-created list of locations does result in a level of selectivity. However, interpolation is a low-level computer vision task requiring only a basic understanding of scene dynamics and the general object shapes. Thus, the list's data diversity is believed to be sufficient. \Cref{tab:vfi_loc_source} reports results on the (full) LAVIB test set when training FLAVR on 700 videos from queries containing only either \texttt{London} 
, \texttt{Istanbul}, or \texttt{Seoul}.

\noindent
\textbf{Potential improvements}. From \Cref{tab:vfi_loc_source}, specific location terms do not show a significant impact on performance. However, including more locations can potentially further increase the variance of some statistics; e.g. ARMS. In addition to weather queries, other terms such as time of day can be added to explicitly enforce diversification in the returned videos.

\subsection{Activities}

\noindent
\textbf{Motivation}. Activity terms are added to avoid static scenes. The distinction between sports and actions subgroups is done to control the expected motion intensity. Activities do however provide a strong constraint for the video content.

\noindent
\textbf{Remedy used}. In total, approximately $\sim$30\% of the queries include actions. The majority of videos returned are either one-shot tours of locations or vlogs. Both types can easily be segmented into 10-second and 1-second clips by the pipeline as they include little to no edits/cuts. Sports are included in a small portion of the queries (~4\%) to avoid specialization. \Cref{tab:vfi_act_source} ablates on 1,000 train videos sourced from queries that include different portions of activity terms. The evaluation is done on the (full) LAVIB test set. The partial inclusion of actions (30\% and 60\%) is shown to be the most balanced strategy for diversity.

\textbf{About}. Two activity categories are defined as motion variances, which present an important challenge in VFI. The \texttt{sports} subgroup primarily includes videos with fast-moving people/objects or camera motion. Specific terms are combined for the following sports; \texttt{climbing}, \texttt{rafting}, \texttt{skiing}, and \texttt{snowboarding} are combined with any of the \{\texttt{forest}, \texttt{mountains}\}, \texttt{snorkeling} is combined with \{\texttt{marinelife}, \texttt{shoreline}, \texttt{underwater}\}, and \texttt{tennis training} is combined with \{\texttt{park}\}. This results in search items such as; `\textcolor{col2}{snowboarding} \textcolor{col4}{mountain} 4K'. The \texttt{action} subgroup is only used in combination with locations.

\noindent
\textbf{Limitations}. Action terms such as \texttt{walking} or \texttt{car ride} are generic and return a large number of videos. Despite viewpoints from hand-held or mounted cameras being some of the most common in online videos, limitations exist.

\noindent
\textbf{Potential improvements}. The videos returned using only location are primarily compilations/highlights from aerial, bird's eye, long shot, or panoramic footages. Driving videos are also ideal for capturing overhead shots. Although both help reduce viewpoint bias, adding an additional vocabulary term based on viewpoint can increase diversity further.

\subsection{Misc}

\begin{figure}[!ht]
    \centering
    \includegraphics[width=\linewidth]{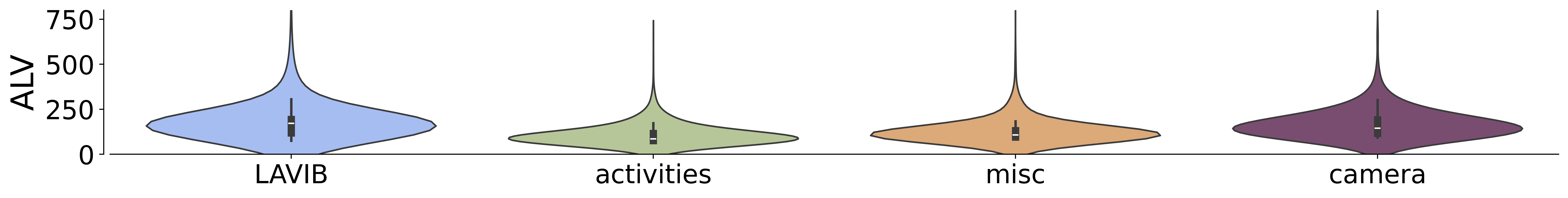}
    \caption{\textbf{ALV distributions} for \textcolor{all}{all LAVIB} and videos from \textcolor{act}{activities}, \textcolor{misc}{misc}, and \textcolor{camera}{camera} queries.}
    \label{fig:alv_tot}
\end{figure}

\begin{figure}[!ht]
    \centering
    \includegraphics[width=\linewidth]{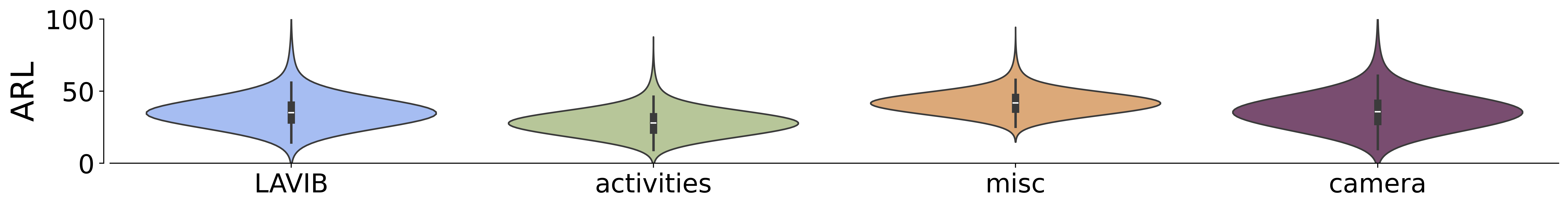}
    \caption{\textbf{ARL distributions} for \textcolor{all}{all LAVIB} and videos from \textcolor{act}{activities}, \textcolor{misc}{misc}, and \textcolor{camera}{camera} queries.}
    \label{fig:arl_tot}
\end{figure}

\noindent
\textbf{Motivation}. Miscellaneous search terms were manually added to diversify the search. The returned videos can vary from the rest of LAVIB by a. different luminance fluctuations; e.g. \texttt{underwater}, b. low; e.g. \texttt{metro} or c. high; e.g. \texttt{birds}, \texttt{flowers}, \texttt{insects}, contrast. 19 camera types are also selected manually to include a variety of phone cameras, action and digital cameras, and DSLRs. The difference in ARL and ALV distributions for misc and camera-based queries compared to the entire LAVIB is shown in \Cref{fig:alv_tot,fig:arl_tot}.

\noindent
\textbf{Remedy used}. Camera terms are added to $\sim$10\% of the queries to avoid returning irrelevant videos. This was done after manually checking the video titles. Approximately 7\% of the dataset is collected with misc terms. \Cref{tab:vfi_misc_source} reports performance on 1,000 train examples that are partially sourced from misc queries. Similarly to \Cref{tab:vfi_act_source}, maintaining a balance between misc and non-misc queries improves generalizability.

\textbf{About}. Miscellaneous search terms are primarily combined with recording equipment to form queries; e.g. `\textcolor{col4}{ocean view} \textcolor{col5}{Yi 4K+} 4K'.

\begin{figure}[!ht]
     \centering
     \begin{subfigure}[b]{0.24\textwidth}
         \centering
         \includegraphics[width=\textwidth]{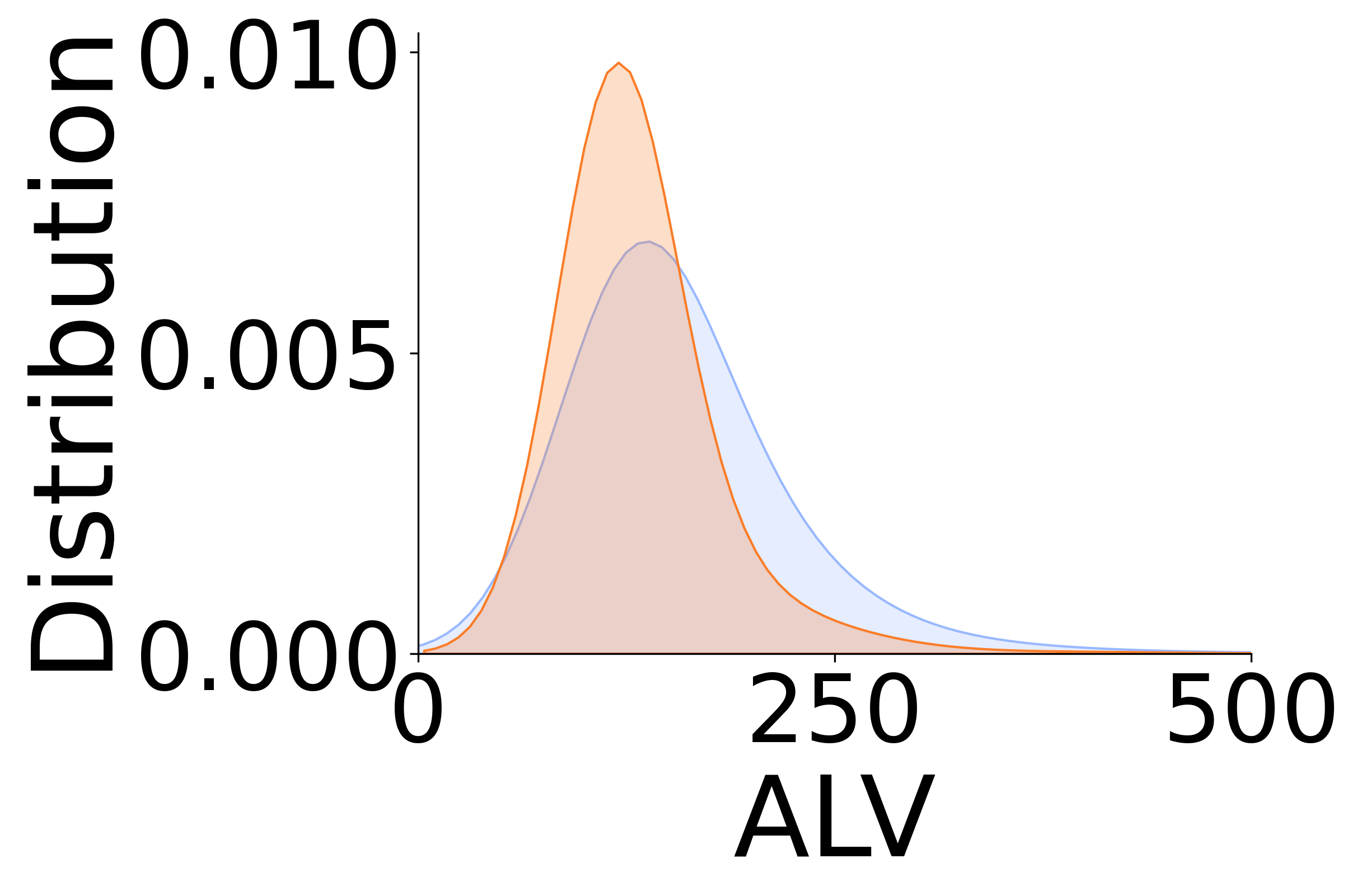}
         \caption{ALV of \textcolor{all}{all LAVIB} and \textcolor{misc}{misc}-only sourced videos.}
         \label{fig:all_misc_alv}
     \end{subfigure}
     \hfill
     \begin{subfigure}[b]{0.24\textwidth}
         \centering
         \includegraphics[width=\textwidth]{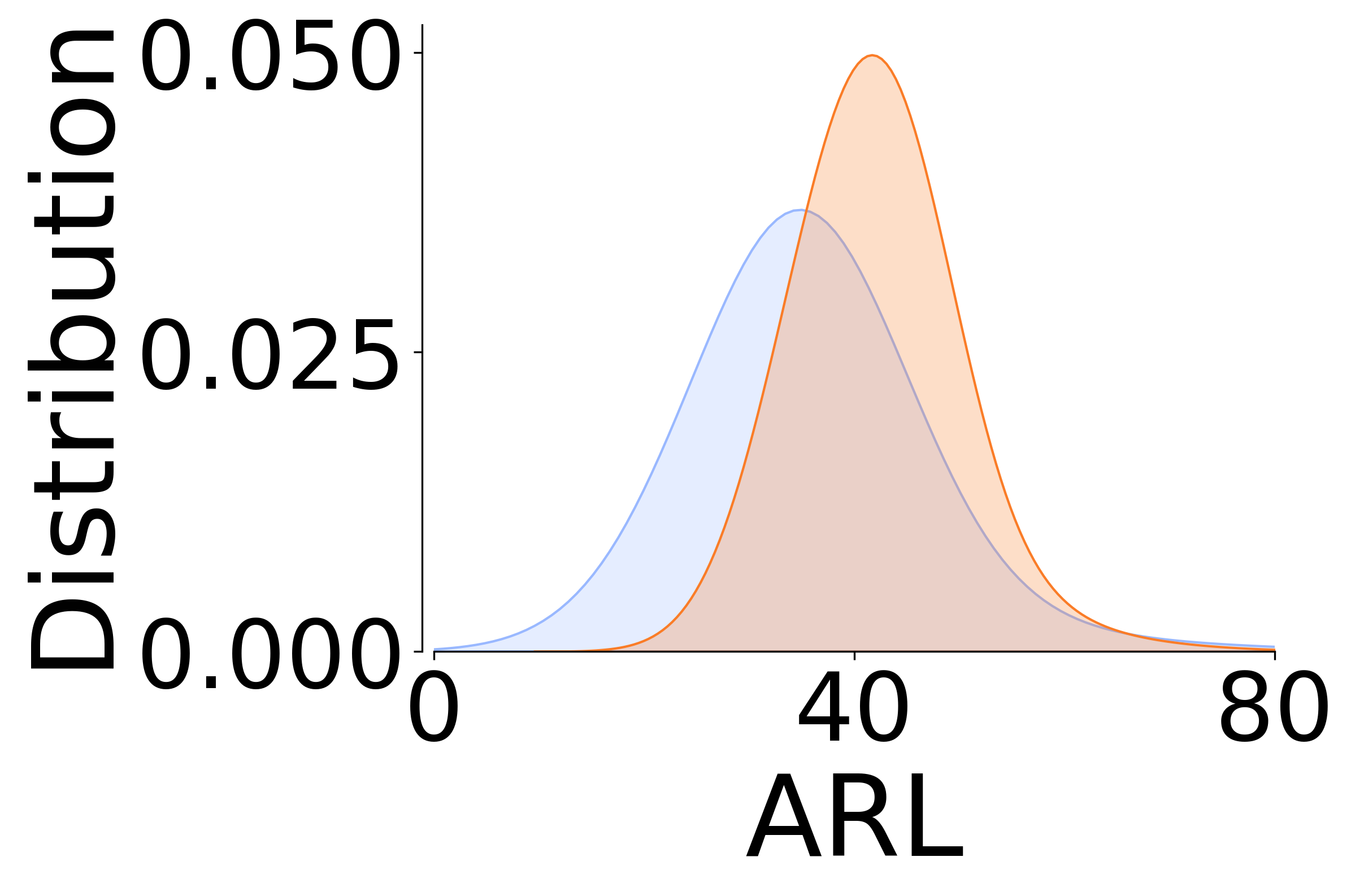}
         \caption{ARL of \textcolor{all}{all LAVIB} and \textcolor{misc}{misc}-only sourced videos.}
         \label{fig:all_misc_arl}
     \end{subfigure}
     \hfill
     \begin{subfigure}[b]{0.24\textwidth}
         \centering
         \includegraphics[width=\textwidth]{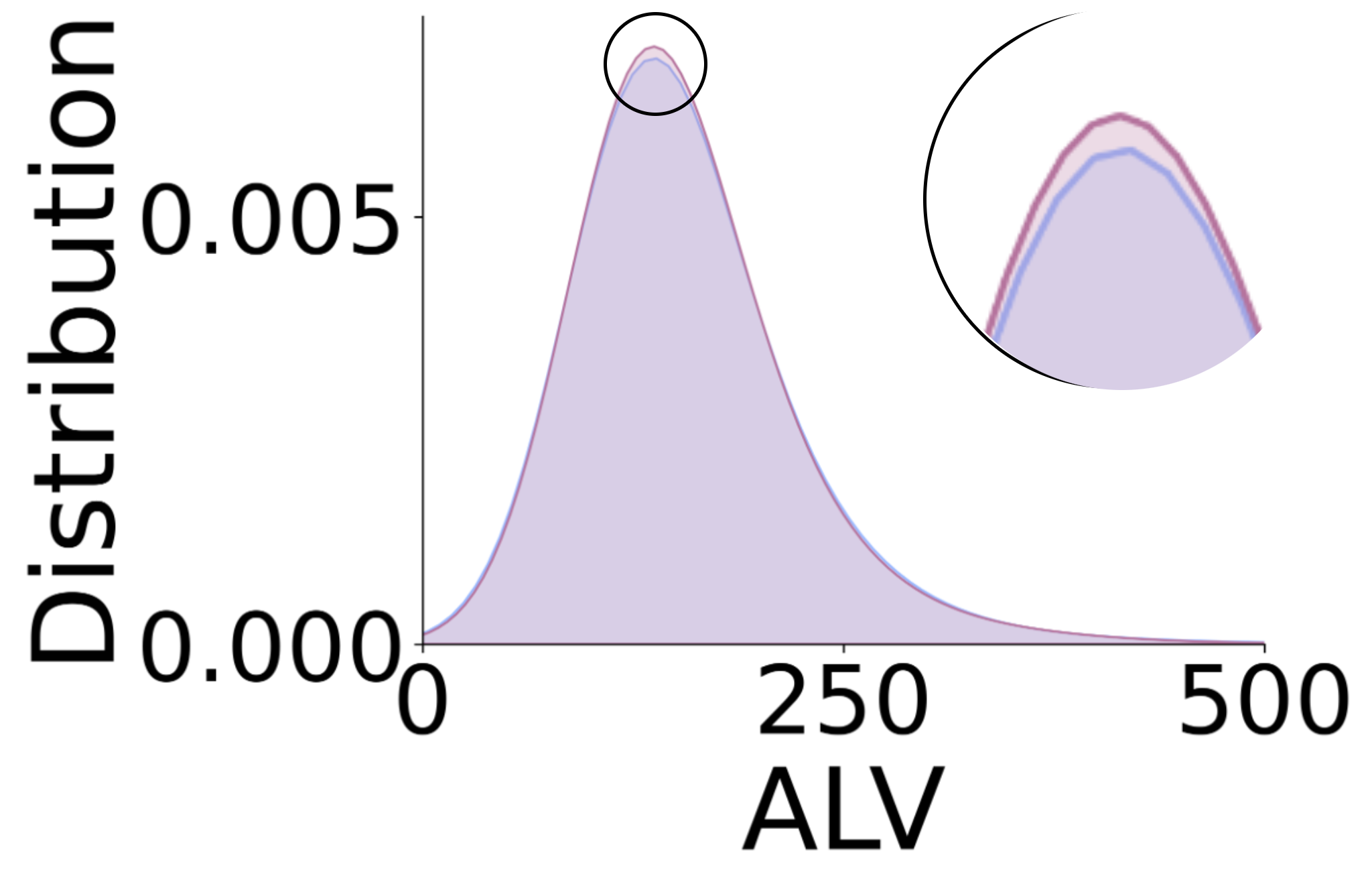}
         \caption{ALV of \textcolor{all}{all LAVIB} and a \textcolor{control}{random subset} of videos.}
         \label{fig:all_random_alv}
     \end{subfigure}
     \begin{subfigure}[b]{0.24\textwidth}
         \centering
         \includegraphics[width=\textwidth]{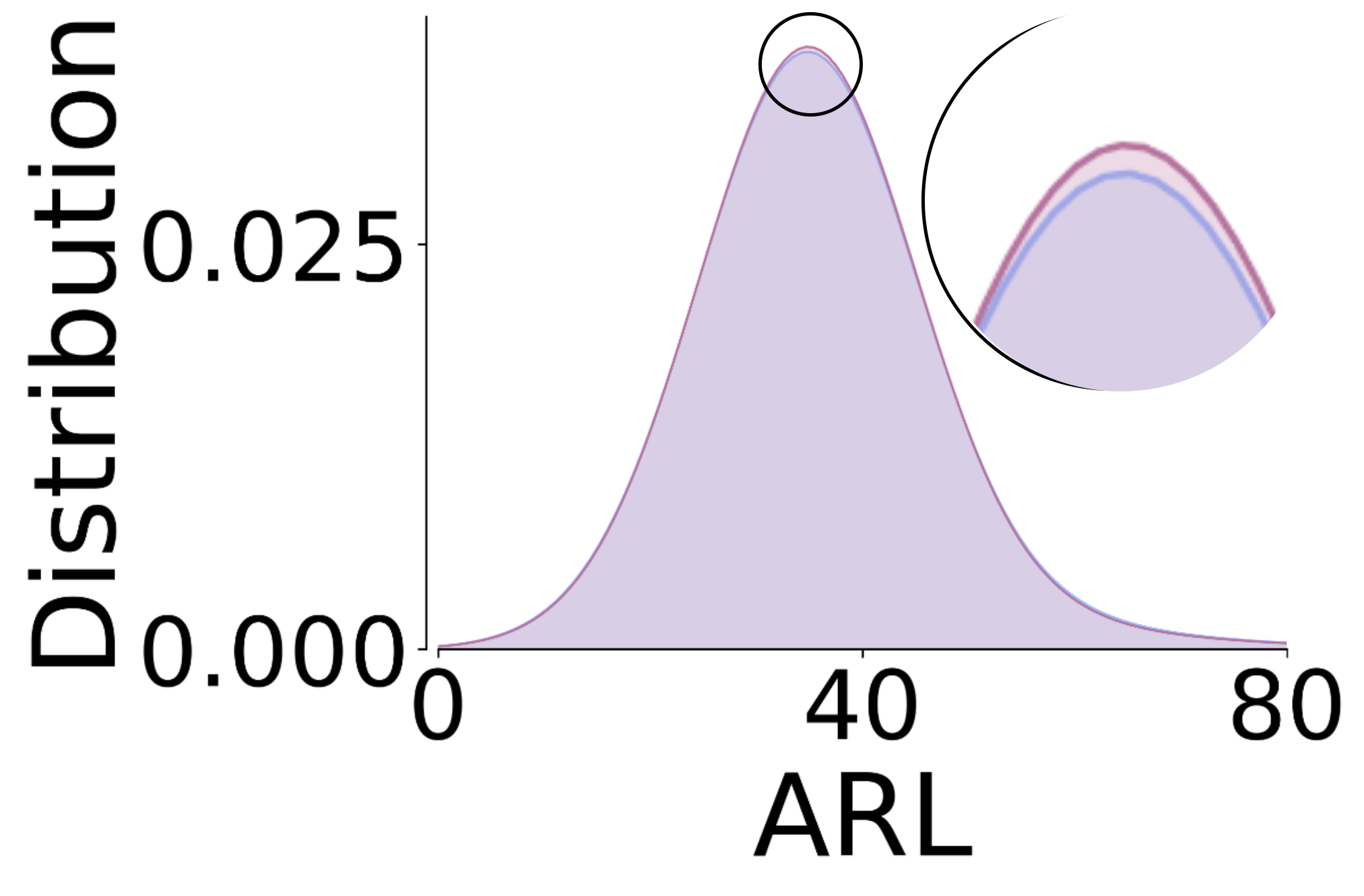}
         \caption{ARL of \textcolor{all}{all LAVIB} and a \textcolor{control}{random subset} of videos.}
         \label{fig:all_random_arl}
     \end{subfigure}
        \caption{\textbf{ARL and ALV distributions} for \textcolor{all}{all LAVIB} videos. Additional distributions from \textcolor{misc}{misc} queries and a \textcolor{control}{random subset}, both of size 19,706 ($\sim$7\% of total videos), are shown for direct comparisons.}
        \label{fig:alv_arl_distributions}
\vspace{-1.2em}
\end{figure}

\noindent
\textbf{Limitations}. The inclusion of misc terms aims to improve diversity. However, as noted, video themes such as screen captures do not guarantee significant variations in video statistics.  The narrow ALV/ARL distributions of videos sourced from misc queries are compared to an equally sized random sample from LAVIB in \Cref{fig:alv_arl_distributions}. Similarly, some camera types may not necessarily differ in video quality.

\noindent
\textbf{Potential improvements}. A further analysis on the misc queries that source videos with the most diverse statistics can highlight the specific terms that improve variance. This can also be used to weigh each term during selection. The same approach can also be applied to the camera types.

\begin{table}[t]
\centering
\begin{minipage}{.325\linewidth}
\caption{\textbf{Results on different location-based train subsets}.}
\resizebox{\linewidth}{!}{%
\setlength\tabcolsep{4.6pt}
\begin{tabular}{l ll}
\toprule
 Term & PSNR$\uparrow$ & SSIM$\uparrow$  \\
 \midrule
 \texttt{London} & 30.69 & 0.945 \\
 \texttt{Istanbul} & 30.75 & 0.944 \\
 \texttt{Seoul} & \textbf{30.81} & \textbf{0.949} \\
\bottomrule
\\
\end{tabular}
\label{tab:vfi_loc_source}
}
\end{minipage}%
\hfill
\begin{minipage}{.32\linewidth}
\caption{\looseness-1 \textbf{Results on different activity-based subsets}.}
\resizebox{\linewidth}{!}{%
\setlength\tabcolsep{4.6pt}
\begin{tabular}{l ll}
\toprule
 Act. (\%) & PSNR$\uparrow$ & SSIM$\uparrow$  \\
 \midrule
 0 & 28.57 & 0.932  \\
 30 & \textbf{31.08} & \textbf{0.953} \\
 60 & 30.65 & 0.948 \\
 100 & 29.23 & 0.940 \\
\bottomrule
\end{tabular}
\label{tab:vfi_act_source}
}
\end{minipage}%
\hfill
\begin{minipage}{.325\linewidth}
\caption{\looseness-1 \textbf{Results on different misc-based subsets}.}
\resizebox{\linewidth}{!}{%
\setlength\tabcolsep{4.6pt}
\begin{tabular}{l ll}
\toprule
 Misc (\%) & PSNR$\uparrow$ & SSIM$\uparrow$  \\
 \midrule
 0 & 29.31 & 0.941 \\
 30 & \textbf{30.54} & \textbf{0.950} \\
 60 & 29.66 & 0.934 \\
 100 & 28.83 & 0.929 \\
\bottomrule
\end{tabular}
\label{tab:vfi_misc_source}
}
\end{minipage}
\vspace{-1.8em}
\end{table}

\begin{algorithm}[t]
  \caption{DUPLEX video selection}\label{alg:duplex}
   \hspace*{\algorithmicindent} \textbf{Input}: dataset $\mathcal{D}$, sets \{ \text{train}, \text{val}, \text{test} \} \\
    \hspace*{\algorithmicindent} \textbf{Output}: dataset splits: $\{\mathcal{D}_{\text{train}},\mathcal{D}_{\text{val}},\mathcal{D}_{\text{test}}\}$ \\
  \begin{algorithmic}[1]
    \For{$\text{set}\_\text{i} \; \in \{ \text{train}, \text{val}, \text{test} \} $}
    \State $\mathbf{s}\leftarrow \text{max}(\{ \| \mathbf{x}_j - \mathbf{x}_k^T \|_2 \})$ where $\mathbf{x}_j,\mathbf{x}_k$ are metrics for videos $j,k$ within $\mathcal{D}$.
    \State $\mathcal{D}_{\text{set}\_\text{i}} \leftarrow \{j,k\}$
    \State $\mathcal{D} \leftarrow \mathcal{D}\setminus \{j,k\}$
    \EndFor
    \For{$\text{set}\_\text{idx} \; \in \{tr,v,ts\}$}
    \For{$i \; \in \{3,\text{set}\_\text{idx}\}$}
        \State $\mathbf{s}_i\leftarrow \text{max}(\{ \| \mathbf{x}_l - \mathcal{D}_{\text{set}\_\text{idx}}[-1]^T \|_2 \})$ where $\mathbf{x}_l$ is a video in $\mathcal{D}$.
        \State $\mathcal{D}_{\text{set}\_\text{idx}} \leftarrow \mathcal{D}_{\text{set}\_\text{idx}} \cup \{\mathbf{x}_l\}$ 
        \State $\mathcal{D} \leftarrow \mathcal{D}\setminus \{\mathbf{x}_l\}$
    \EndFor
    \EndFor
  \end{algorithmic}
\end{algorithm}

\section{Video sorting}
\label{app:sort}

For the benchmark, each split should have similar video metric distributions. Due to the multi-dimensionality and high variance across metrics,  
DUPLEX is used to calibrate dataset split sampling. DUPLEX uses the L2 distance across video metrics when creating train/val/test splits. For each set, the algorithm discovers the two most distant videos given their AFM, ALV, ARMS, and ARL metrics. It then iteratively samples videos that maximize the distance to previously sampled videos. This is done iteratively until the size condition for the split is met. \Cref{alg:duplex} provides a programmatic view of DUPLEX sampling.

\section{Detailed training settings}
\label{app:settings}

All training experiments are done with the codebases provided by the authors with $2\times$ Nvidia L40 with an average training time of 2 days per model. Computational settings for each model are reported in \Cref{tab:rife_params,tab:emavfi_params,tab:flavr_params}.

\begin{table}[t]
\centering
\begin{minipage}{.325\linewidth}
\caption{\textbf{RIFE settings}}
\resizebox{\linewidth}{!}{%
\setlength\tabcolsep{1.3pt}
\begin{tabular}{l l}
\toprule
 Parameter & value \\
 \midrule
 \texttt{batch size} & 64 \\
 \texttt{optimizer} & AdamW \\
 \texttt{weight decay} & 1e$^{\!\!-6}$ \\
 \texttt{learning rate} & 1e$^{\!\!-4}$ \\
 \texttt{learning scheduler} & \texttt{Step} \\
 \texttt{additional params} & \makecell{$\!\!\!$\texttt{beta1}=0.9 \\ \texttt{beta2}=0.99} \\
 
\bottomrule
\end{tabular}
\label{tab:rife_params}
}
\end{minipage}%
\hfill
\begin{minipage}{.325\linewidth}
\caption{\textbf{EMA-VFI settings}.}
\resizebox{\linewidth}{!}{%
\setlength\tabcolsep{1.3pt}
\begin{tabular}{l l}
\toprule
 Parameter & value \\
 \midrule
 \texttt{batch size} & 64 \\
 \texttt{optimizer} & AdamW \\
 \texttt{weight decay} & 1e$^{\!\!-4}$ \\
 \texttt{learning rate} & 1e$^{\!\!-4}$ \\
 \texttt{learning scheduler} & \texttt{Warmup}\\
 \texttt{additional params} &\makecell{$\!\!\!$\texttt{beta1}=0.9 \\ 
 \texttt{beta2}=0.99} \\
 
\bottomrule
\end{tabular}
\label{tab:emavfi_params}
}
\end{minipage}%
\hfill
\begin{minipage}{.325\linewidth}
\caption{\textbf{FLAVR settings}}
\resizebox{\linewidth}{!}{%
\setlength\tabcolsep{1.3pt}
\begin{tabular}{l l}
\toprule
 Parameter & value \\
 \midrule
 \texttt{batch size} & 64 \\
 \texttt{optimizer} & Adam \\
 \texttt{weight decay} & 1e$^{\!\!-6}$ \\
 \texttt{learning rate} & 5e$^{\!\!-3}$ \\
 \texttt{learning scheduler} & \texttt{Step} \\
 \texttt{additional params} & \makecell{$\!\!\!$\texttt{beta1}=0.9 \\ \texttt{beta2}=0.99} \\

\bottomrule
\end{tabular}
\label{tab:flavr_params}
}
\end{minipage}
\vspace{-1.2em}
\end{table}

\section{Ablations}
\label{app:ablations}

Supplementary to the main results in \S\ref{sec:results} ablations are performed with FLAVR for variations in train set sizes for both benchmark and OOD challenges. 

\begin{figure}[!t]
\begin{minipage}[T]{\textwidth}
\centering
\vspace{.5em}
\begin{minipage}[t]{.65\linewidth}
\captionof{table}{\looseness-1 \textbf{Val and test set results} when training on different portions of the train set. \textit{full} denotes that the entire train set from LAVIB is retained for training. Best results per split are in \textbf{bold}.}
\resizebox{\linewidth}{!}{%
\begin{tabular}{l lll lll}
\toprule
    \multirow{2}{*}{\makecell{LAVIB\\ train \%}} & \multicolumn{3}{c}{val set} & \multicolumn{3}{c}{test set} \\
  & PSNR$\uparrow$ & SSIM$\uparrow$ & LPIPS$\downarrow$ & PSNR$\uparrow$ & SSIM$\uparrow$ & LPIPS$\downarrow$ \\
 \midrule
 20\% & 29.43 & 0.895 & 8.257e$^{\!\!-2}$ & 29.48 & 0.894 & 8.472e$^{\!\!-2}$ \\
  40\% & 31.68 & 0.960 & 4.108e$^{\!\!-2}$ & 31.63 & 0.958 & 4.241e$^{\!\!-2}$\\
  60\% & 32.64 & 0.970 & 3.566e$^{\!\!-2}$ & 32.50 & 0.967 & 3.835e$^{\!\!-2}$ \\
  80\% & 33.36 & 0.975 & 2.971e$^{\!\!-2}$  & 33.19 & 0.973 & 3.064e$^{\!\!-2}$ \\
  \rowcolor{LightGrey} \textit{full} & \textbf{33.72} & \textbf{0.981} & \textbf{2.515e}$^{\!\!-2}$ & \textbf{33.44} & \textbf{0.981} & \textbf{2.934e}$^{\!\!-2}$ \\
\bottomrule
\end{tabular}
\label{tab:vfi_partial_test}
}
\end{minipage}%
\hfill
\begin{minipage}[t]{.325\linewidth}
\vspace{.5em}
\includegraphics[width=\linewidth,valign=t]{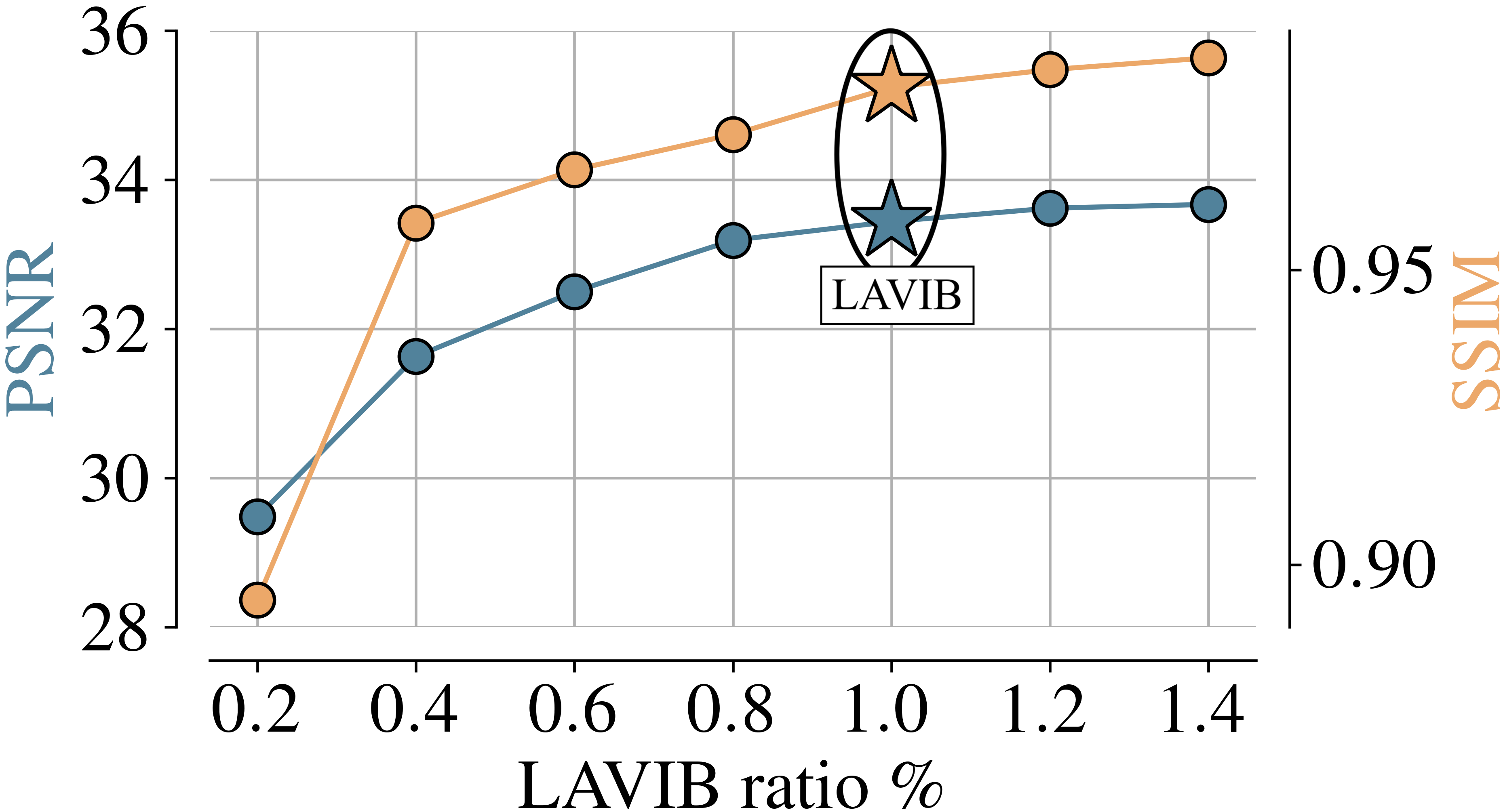}
\captionof{figure}{\textbf{\looseness-1 Test PSNR/SSIM over train sizes}. In ratios $<1.0\%$ clips are removed. In ratios $>1.0\%$ clips are added from left-out segments.}
\label{fig:var}
\end{minipage}%
\end{minipage}
\end{figure}

\textbf{Benchmarks over reduced training set sizes}.
\Cref{tab:vfi_partial_test} presents val and test set results with reductions in the training set sizes. At each reduction setting, clips are dropped randomly. Performance drops significantly for both validation and test sets as the size of the training set decreases with an average -4.12 and -0.086 PSNR/SSIM.

\begin{table}[ht]
\begin{minipage}{.49\linewidth}
\caption{\textbf{AFM OOD ablation results}.}
    \centering
    \resizebox{\linewidth}{!}{%
    \setlength\tabcolsep{4.5pt}
    \begin{tabular}{clccc}
    \toprule
    AFM sett. & Train \% removed & PSNR$\uparrow$ & SSIM$\uparrow$ & LPIPS$\downarrow$ \\ 
    \midrule
         \multirow{3}{*}{\motionslow{}\raisebox{.2em}{$\rightarrow$}\motionfast{}} & - bottom 30\% & 29.45 & 0.912 & 5.637e$^{\!\!-2}$   \\
         & - top 30 \% & 26.32 & 0.854 & 9.428e$^{\!\!-2}$ \\
         & \cellcolor{LightGrey} \textit{None} & \cellcolor{LightGrey} \textbf{30.67} & \cellcolor{LightGrey} \textbf{0.959} & \cellcolor{LightGrey} \textbf{5.094e}$^{\!\!-2}$ \\
         \midrule
         \multirow{3}{*}{\motionfast{}\raisebox{.2em}{$\rightarrow$}\motionslow{}} & - bottom 30\%  & 32.80 & 0.973 & 3.396e$^{\!\!-2}$  \\
         & - top 30 \% & 35.32 & 0.987 & 1.503e$^{\!\!-2}$  \\
         & \cellcolor{LightGrey} \textit{None} & \cellcolor{LightGrey} \textbf{35.66} & \cellcolor{LightGrey} \textbf{0.991} & \cellcolor{LightGrey} \textbf{1.342e}$^{\!\!-2}$\\
    \bottomrule
    \end{tabular}
    }
    \label{tab:var_afm}
\end{minipage}%
\hfill
\begin{minipage}{.49\linewidth}
\caption{\textbf{ALV OOD ablation results}.}
    \centering
    \resizebox{\linewidth}{!}{%
    \setlength\tabcolsep{4.5pt}
    \begin{tabular}{clccc}
    \toprule
    ALV sett. & Train \% removed & PSNR$\uparrow$ & SSIM$\uparrow$ & LPIPS$\downarrow$ \\ 
    \midrule
         \multirow{3}{*}{\blurlow{}\raisebox{.2em}{$\rightarrow$}\blurhigh{}} & - bottom 30\% & 30.32 & 0.933 & 4.781e$^{\!\!-2}$   \\
         & - top 30 \% & 28.54 & 0.905 & 5.567e$^{\!\!-2}$ \\
         & \cellcolor{LightGrey} \textit{None} & \cellcolor{LightGrey} \textbf{31.78} & \cellcolor{LightGrey} \textbf{0.962} & \cellcolor{LightGrey} \textbf{2.942e}$^{\!\!-2}$ \\
         \midrule
         \multirow{3}{*}{\blurhigh{}\raisebox{.2em}{$\rightarrow$}\blurlow{}} & - bottom 30\% & 31.24 & 0.958 & 4.396e$^{\!\!-2}$   \\
         & - top 30 \% & 34.35 & 0.971 & 2.740e$^{\!\!-2}$ \\
         & \cellcolor{LightGrey} \textit{None} & \cellcolor{LightGrey} \textbf{34.67} & \cellcolor{LightGrey} \textbf{0.975} & \cellcolor{LightGrey} \textbf{2.627e}$^{\!\!-2}$ \\
    \bottomrule
    \end{tabular}
    }
    \label{tab:var_alv}
\end{minipage}\\
\begin{minipage}{.49\linewidth}
\caption{\textbf{ARMS OOD ablation results}.}
    \centering
    \resizebox{\linewidth}{!}{%
    \setlength\tabcolsep{4.5pt}
    \begin{tabular}{clccc}
    \toprule
    ARMS sett. & Train \% removed & PSNR$\uparrow$ & SSIM$\uparrow$ & LPIPS$\downarrow$ \\ 
    \midrule
         \multirow{3}{*}{\contrastlow{}\raisebox{.2em}{$\rightarrow$}\contrasthigh{}} & - bottom 30\% & 32.87 & 0.977 & 2.683e$^{\!\!-2}$    \\
         & - top 30 \% & 31.92 & 0.965 & 3.769e$^{\!\!-2}$  \\
         & \cellcolor{LightGrey} \textit{None} & \cellcolor{LightGrey} \textbf{33.02} & \cellcolor{LightGrey} \textbf{0.982} & \cellcolor{LightGrey} \textbf{2.561e}$^{\!\!-2}$ \\
         \midrule
         \multirow{3}{*}{\contrasthigh{}\raisebox{.2em}{$\rightarrow$}\contrastlow{}} & - bottom 30\%  & 30.18 & 0.931 & 4.515e$^{\!\!-2}$  \\
         & - top 30 \% & 30.74 & 0.973 & 3.327e$^{\!\!-2}$ \\
         & \cellcolor{LightGrey} \textit{None} & \cellcolor{LightGrey} \textbf{31.11} & \cellcolor{LightGrey} \textbf{0.977} & \cellcolor{LightGrey} \textbf{3.024e}$^{\!\!-2}$ \\
    \bottomrule
    \end{tabular}
    }
    \label{tab:var_arms}
\end{minipage}%
\hfill
\begin{minipage}{.49\linewidth}
\caption{\textbf{ARL OOD ablation results}.}
    \centering
    \resizebox{\linewidth}{!}{%
    \setlength\tabcolsep{4.5pt}
    \begin{tabular}{clccc}
    \toprule
    ARL sett. & Train \% removed & PSNR$\uparrow$ & SSIM$\uparrow$ & LPIPS$\downarrow$ \\ 
    \midrule
         \multirow{3}{*}{\lumlow{}\raisebox{.2em}{$\rightarrow$}\lumhigh{}} & - bottom 30\% & 32.76 & 0.973 & 2.806e$^{\!\!-2}$  \\
         & - top 30 \% & 32.15 & 0.961 & 3.315e$^{\!\!-2}$  \\
         & \cellcolor{LightGrey} \textit{None} & \cellcolor{LightGrey} \textbf{33.97} & \cellcolor{LightGrey} \textbf{0.980} & \cellcolor{LightGrey} \textbf{2.543e}$^{\!\!-2}$ \\
         \midrule
         \multirow{3}{*}{\lumhigh{}\raisebox{.2em}{$\rightarrow$}\lumlow{}} & - bottom 30\% & 34.06 & 0.972 & 2.763e$^{\!\!-2}$ \\
         & - top 30 \% & 33.67 & 0.970 & 3.457e$^{\!\!-2}$ \\
         & \cellcolor{LightGrey} \textit{None} & \cellcolor{LightGrey} \textbf{34.20} & \cellcolor{LightGrey} \textbf{0.976} & \cellcolor{LightGrey}  \textbf{2.875e}$^{\!\!-2}$ \\
    \bottomrule
    \end{tabular}
    }
    \label{tab:var_arl}
\end{minipage} \\
\end{table}

\textbf{Performance over varying size}. Motivated by the performance reductions observed with decreases in the train set size in \Cref{tab:vfi_partial_test}, \Cref{fig:var} presents PSNR/SSIM performance when an additional number of clips is retained during the selection progress. Clips are added by relaxing the threshold values. Although the performance improvements observed when including more clips in training in small ratios are significant, this is not retraced with further increases in the size of the current dataset. This shows that the selection process for LAVIB enables the creation of a diverse dataset.

\textbf{OOD over reduced train set sizes}. Performance trends when removing the highest/lowest valued clips in OOD challenges given their metrics are reported in \Cref{tab:var_afm} and \Cref{tab:var_alv} for AFM and ALV. Similarly, \Cref{tab:var_arms} and \Cref{tab:var_arl} report results with training set reductions for ARMS and ARL. Across settings, the portions closer to the target domain; e.g. the top 30\% for the low to high settings and bottom 30\% for the high to low settings present the largest drop in performance when removed. In contrast, when portions of the data that are less similar to the target domain are removed the reductions in performance are marginal. This shows that VFI method trained on domain-specific videos cannot generalize as effectively.

\section{Qualitative}
\label{app:qualitative}

\begin{figure}[t]
    \centering
    \includegraphics[width=\linewidth]{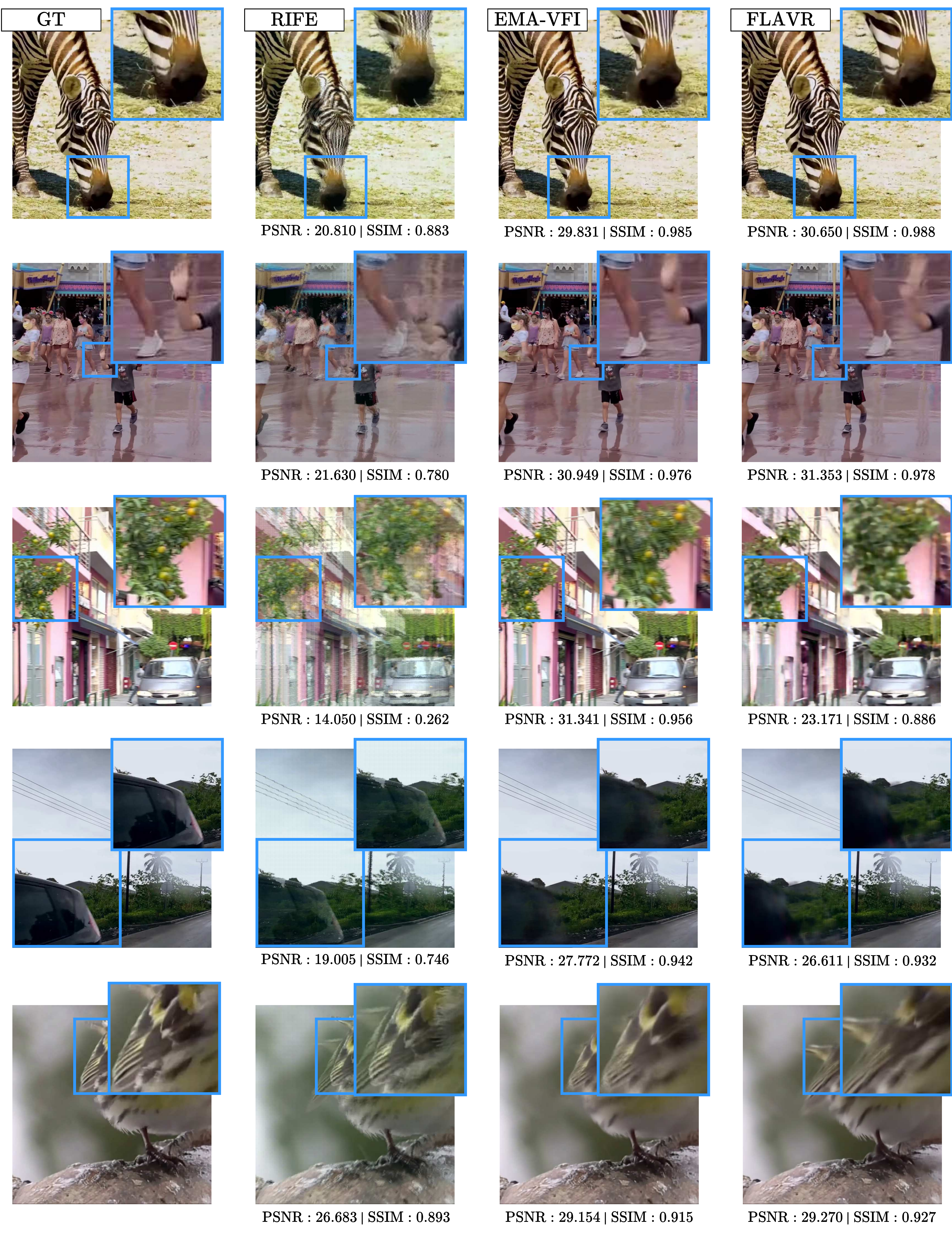}
    \caption{\textbf{Examples from the LAVIB benchmark} (best viewed digitally)} 
    \label{fig:qualitative_main}
    \vspace{-1.2em}
\end{figure}

\begin{figure}[t]
    \centering
    \includegraphics[width=\linewidth]{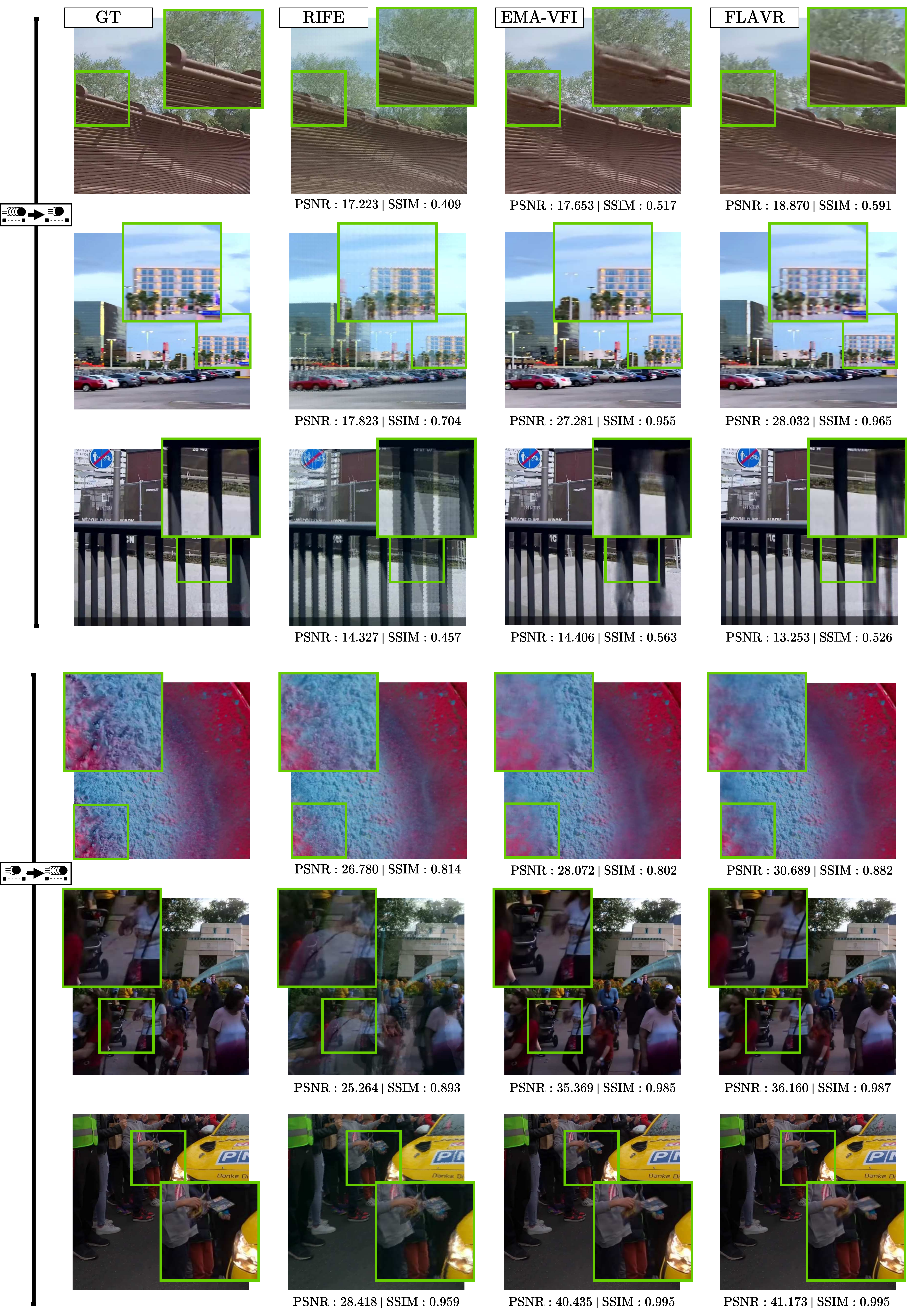}
    \caption{\textbf{Examples of AFM OOD challenges} (best viewed digitally)} 
    \label{fig:qualitative_afm}
    \vspace{-1.2em}
\end{figure}

\begin{figure}[t]
    \centering
    \includegraphics[width=\linewidth]{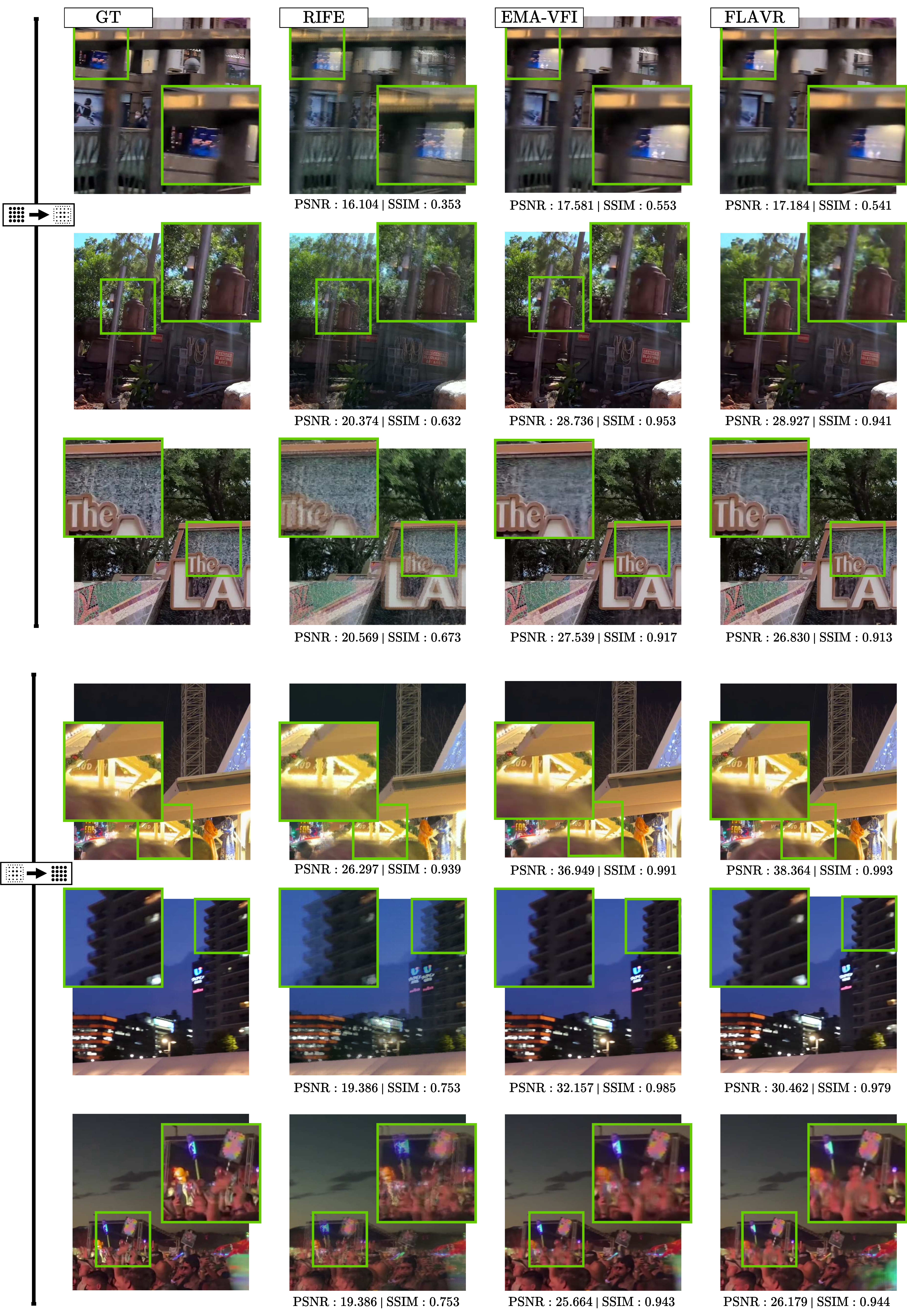}
    \caption{\textbf{Examples of ALV OOD challenges} (best viewed digitally)} 
    \label{fig:qualitative_alv}
    \vspace{-1.2em}
\end{figure}

\begin{figure}[t]
    \centering
    \includegraphics[width=\linewidth]{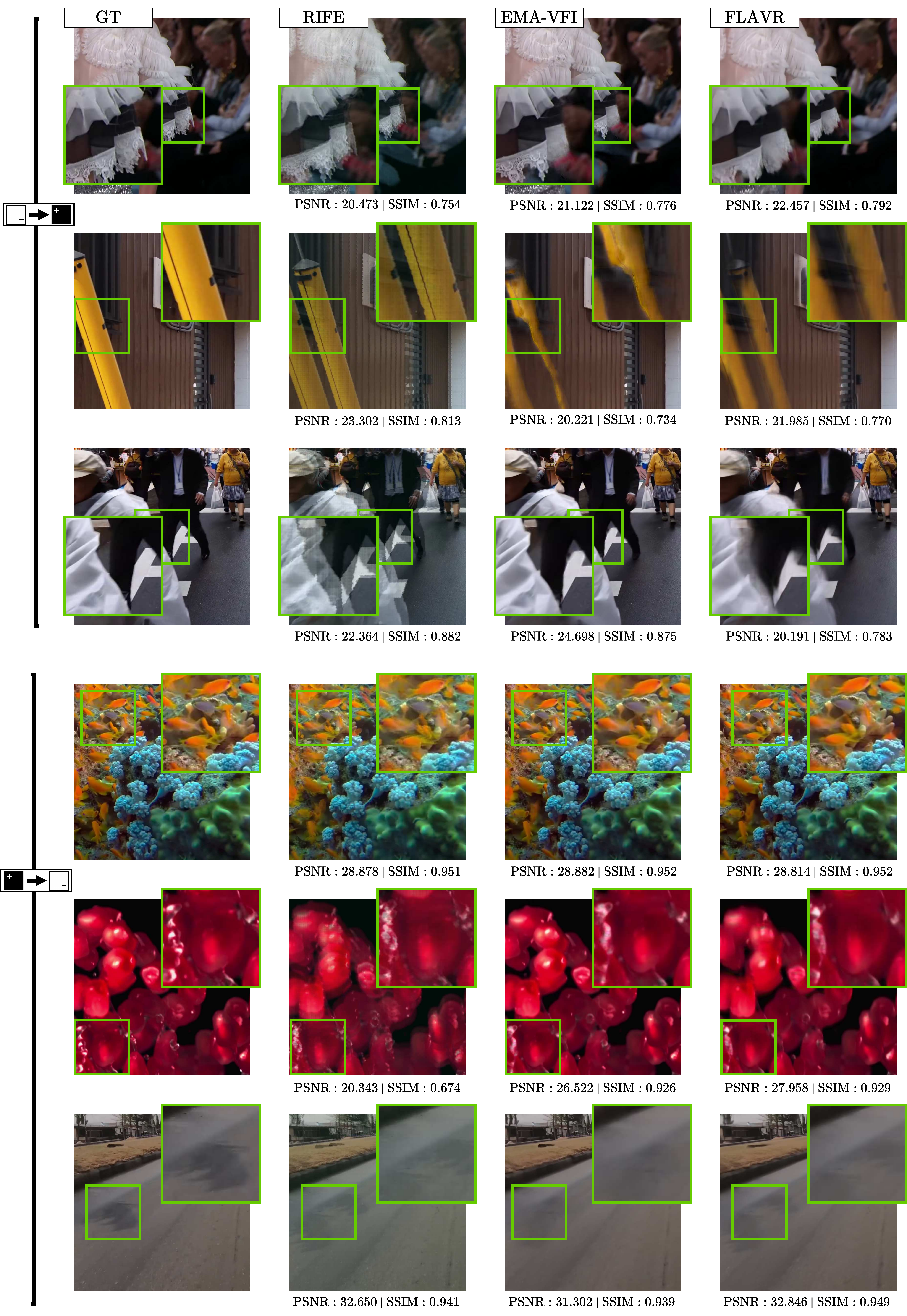}
    \caption{\textbf{Examples of ARMS OOD challenges} (best viewed digitally)} 
    \label{fig:qualitative_arms}
    \vspace{-1.2em}
\end{figure}

\begin{figure}[t]
    \centering
    \includegraphics[width=\linewidth]{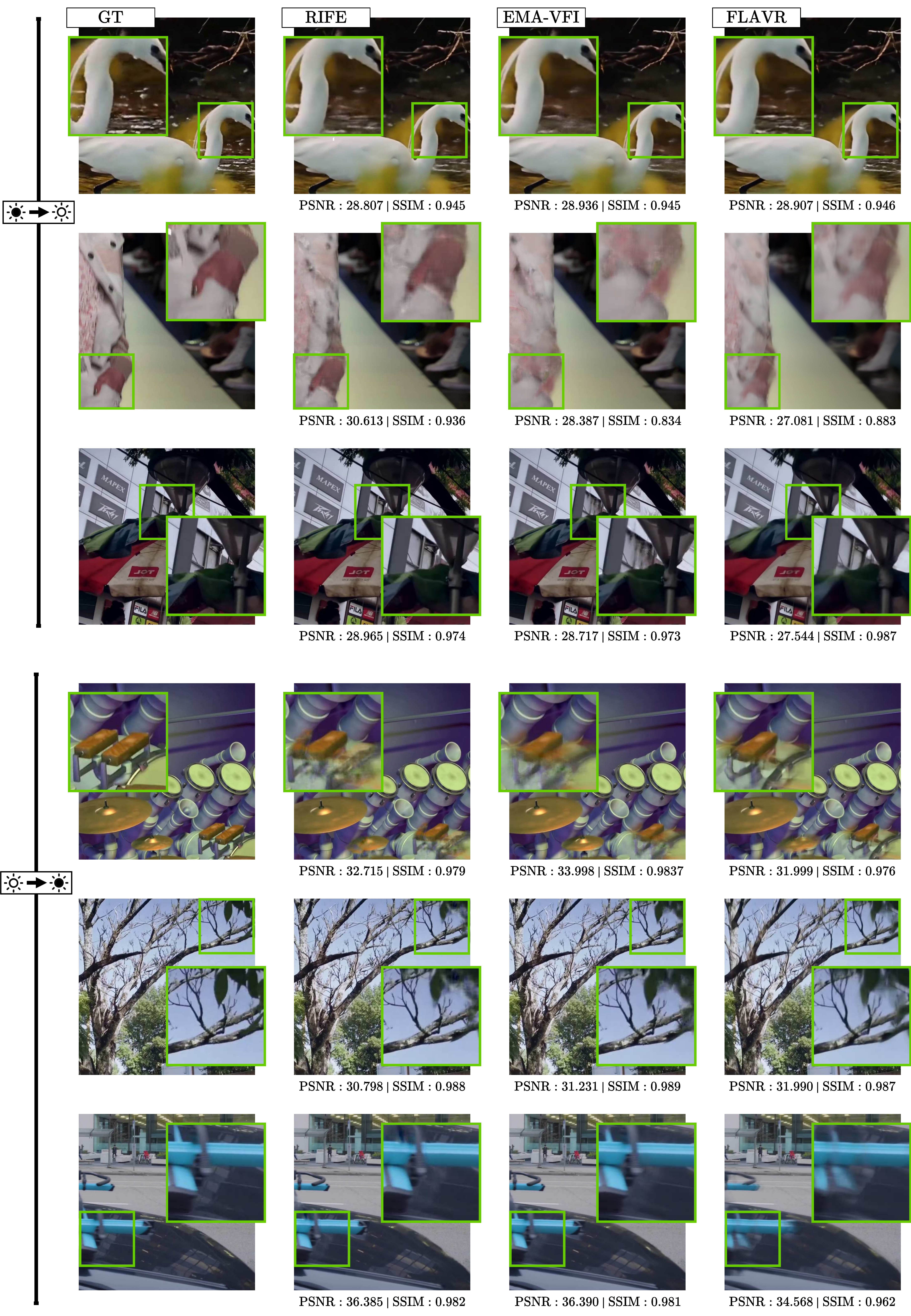}
    \caption{\textbf{Examples of ARL OOD challenges} (best viewed digitally)} 
    \label{fig:qualitative_arl}
    \vspace{-1.2em}
\end{figure}

\Cref{fig:qualitative_main} presents predicted frames from each model on examples from the benchmark test set. Interpolated frames for AFM-, ALV-, ARMS-, and ARL-based OOD challenges are shown in \Cref{fig:qualitative_afm,fig:qualitative_alv,fig:qualitative_arms,fig:qualitative_arl}. In all settings, models can only partially interpolate the unseen frames. The majority of the errors observed are related to high-motion low-contrast examples. In instances where motion blur is present in the ground truth; e.g row 2 \Cref{fig:qualitative_main}, row 5 in \Cref{fig:qualitative_afm}, and rows 1,5, and 6 in \Cref{fig:qualitative_alv}, motion blur is exacerbated at the interpolated frames from all models. Models trained on settings where fine details are not visible such as low ARMS and low ARL only interpolate the general shapes of objects and structures as shown in rows 1 and 2 in \Cref{fig:qualitative_arms} and rows 1--3 in \Cref{fig:qualitative_arl}.

\section{Ethics, privacy, and use}
\label{app:ethics}

\textbf{Ethics and privacy}. The introduced dataset primarily considers footage of landscapes, objects, nature, animals, and screen recordings. However, certain videos may include people. Scenes in which people appear are characterized by high camera motion, scene clutter, and partial visibility of faces that appear briefly for a few seconds. Thus, it is believed that the risk of identification is low. In addition, the video segments from which the dataset is sourced are 1 second long, limiting the number of frames available. As videos are sourced from YouTube a list of the links to the original videos is also provided. 

\textbf{Use}. The dataset is distributed for open-source scientific projects under a Creative Common's Attribution-NonCommercial-Share-Alike (CC BY-SA-NC 4.0). The dataset can be further shared, and adapted, but cannot be used for commercial applications. Adaptations or sharing of the dataset needs to be done under the same license$^{\textcolor{red}{\dagger}}$. \footnotetext{$^{\textcolor{red}{\dagger}}$Clarifications on special use cases can be found in: \url{https://creativecommons.org/licenses/by-nc-sa/4.0/deed.en}}  

\end{document}